\documentclass[10pt,twocolumn,letterpaper]{article}
\usepackage{cvpr}
\usepackage{times}
\usepackage{epsfig}
\usepackage{graphicx}
\usepackage{amsmath}
\usepackage{amssymb}
\usepackage{xcolor}
\usepackage{enumerate}
\usepackage{enumitem}
\usepackage[olditem,oldenum]{paralist}
\usepackage{csquotes}
\usepackage{caption}
\usepackage{subcaption}
\usepackage{verbatim}
\usepackage{color}
\usepackage{fixltx2e}
\usepackage[pagebackref=true,breaklinks=true,letterpaper=true,colorlinks,bookmarks=false,citecolor=green,linkcolor=red]{hyperref}
\usepackage[normalem]{ulem}  
\usepackage{mysymbols}
\usepackage{multirow}
\usepackage{rotating}
\usepackage{booktabs}
\def\funniness{\textit{F\textsubscript{i}}}
\definecolor{gray}{RGB}{0.66, 0.66, 0.66}
\definecolor{orange}{RGB}{225, 90, 0}
\definecolor{teal}{RGB}{5, 210, 150}
\definecolor{yellow}{RGB}{220, 210, 10}
\definecolor{purple}{RGB}{100, 0, 205}
\newcommand{\change}{\textcolor{red}}

\cvprfinalcopy 
\def\cvprPaperID{200} 

\ifcvprfinal\fi
\begin{document}
%
\title{We Are Humor Beings: Understanding and Predicting Visual Humor}
\vspace{0.05\textwidth}\author{Arjun Chandrasekaran$^1$ \quad Ashwin K. Vijayakumar$^1$ \quad Stanislaw Antol$^1$ \quad Mohit Bansal$^2$ \\ Dhruv Batra$^1$ \quad C. Lawrence Zitnick$^3$ \quad Devi Parikh$^1$\vspace{0.005\textwidth}\\
$^1${\fontsize{11}{12}\selectfont Virginia Tech}\quad $^2${\fontsize{11}{12}\selectfont TTI-Chicago}\quad $^3${\fontsize{11}{12}\selectfont Facebook AI Research} \\
{\tt\small $^1$\{carjun, ashwinkv, santol, dbatra, parikh\}@vt.edu} \enspace {\tt\small $^2$mbansal@ttic.edu} \enspace {\tt\small $^3$zitnick@fb.com} }
%
\vspace{-0.02\textwidth}
\maketitle
\thispagestyle{empty}
\begin{abstract}
\vspace{-0.02\textwidth}
Humor is an integral part of human lives. Despite being tremendously impactful, it is perhaps surprising that we do not have a detailed understanding of humor yet. As interactions between humans and AI systems increase, it is imperative that these systems are taught to understand subtleties of human expressions such as humor. In this work, we are interested in the question -- what content in a scene causes it to be funny? As a first step towards understanding visual humor, we analyze the humor manifested in abstract scenes and design computational models for them. We collect two datasets of abstract scenes that facilitate the study of humor at both the scene-level and the object-level. We analyze the funny scenes and explore the different types of humor depicted in them via human studies.
We model two tasks that we believe demonstrate an understanding of some aspects of visual humor. The tasks involve predicting the funniness of a scene and altering the funniness of a scene.
We show that our models perform well quantitatively, and qualitatively through human studies. Our datasets are publicly available.
\end{abstract}
\vspace{-0.03\textwidth} 
\section{Introduction}
\label{sec:intro}
\vspace{-0.005\textwidth}
An adult laughs 18 times a day \cite{DailyOccurenceLaughter} on average. A good sense of humor is related to communication competence \cite{CommunicativeAdaptability1, CommunicativeAdaptability2}, helps raise an individual's social status \cite{EmotionalStates}, popularity \cite{OnBeingWitty, ChildrenSocialDev}, and helps attract compatible mates \cite{AppreciationHumorSexuallySelectedTraits, EvolutionHumanIntrasexualCompetition, HumorInterpersonalAttraction}. 
Humor in the workplace improves camaraderie and helps workers cope with daily stresses \cite{humor_at_work} and loneliness \cite{HumorOrientationLoneliness}.
\emph{fMRI} \cite{fMRI} studies of the brain reveal that humor activates the components of the brain that are involved in reward processing~\cite{BrainActivation}. This probably explains why we actively seek to experience and create humor \cite{HumorRewardRegions}. \par
Despite the tremendous impact that humor has on our lives, the lack of a rigorous definition of humor has hindered humor-related research in the past \cite{LinguisticTheoriesOfHumor, LaStrutturaDellaParodia}. While verbal humor is better understood today \cite{GVTH, SSTH}, visual humor remains unexplored. As vision and AI researchers we are interested in the following question -- what content in an image causes it to be funny? Our work takes a step in the direction of building computational models for visual humor.
\noindent
\begin{figure}[t]
\setlength{\fboxsep}{0pt}
\setlength{\fboxrule}{0pt}
\centering
	\begin{subfigure}[t]{1.5in}
		\fbox{\includegraphics[width=1.67in,bb=0 0 700 400]{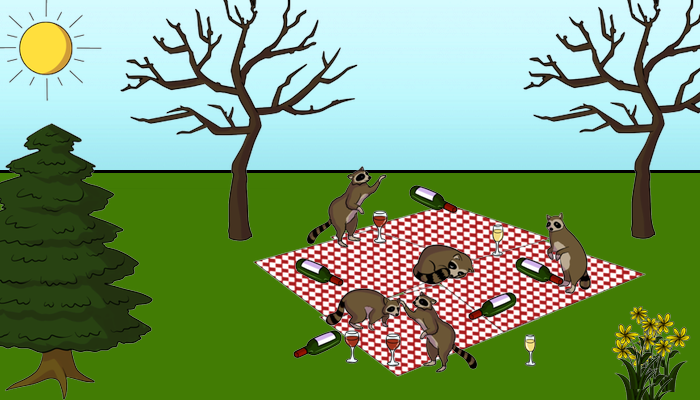}}
		\vspace{-17pt}
		\caption{\emph{Funny scene:} Raccoons are drunk at a picnic.}\label{fig:raccoon_party}
	\end{subfigure}
	\quad
	\begin{subfigure}[t]{1.5in}
		\fbox{\includegraphics[width=1.67in]{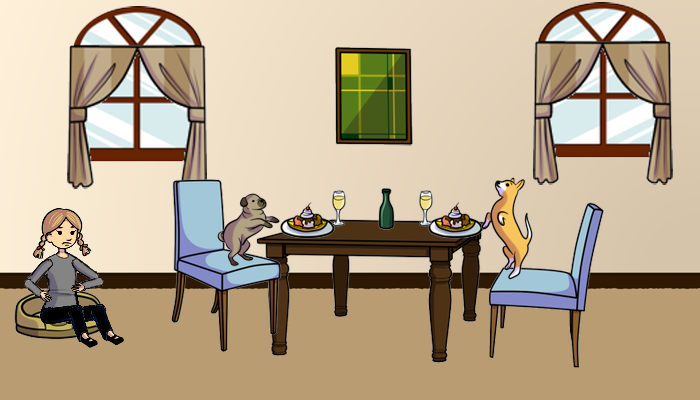}}
		\vspace{-17pt}
		\caption{\emph{Funny scene:} Dogs feast 
		while the girl sits in a pet bed.}\label{fig:dog_dinner}
	\end{subfigure}
	\quad
	\begin{subfigure}[t]{1.5in}
		\fbox{\includegraphics[width=1.67in]{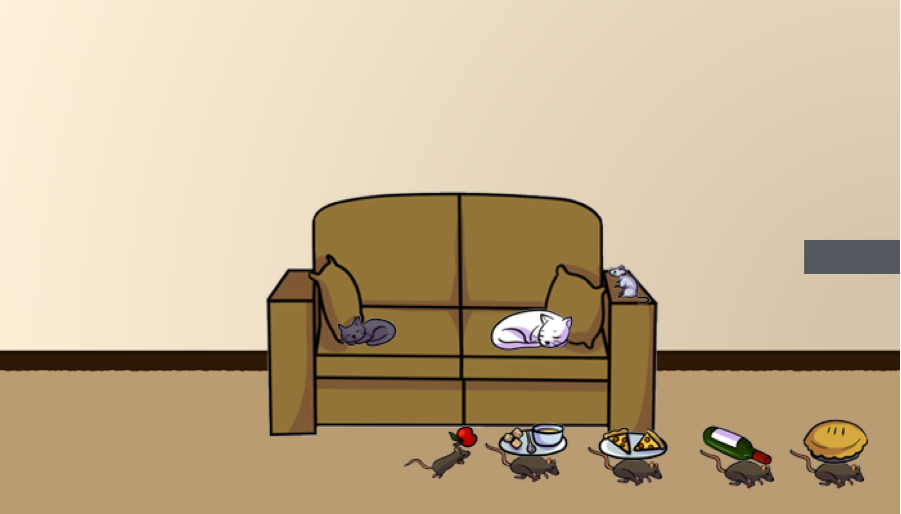}}
		\vspace{-17pt}
		\caption{\emph{Funny scene:} Rats steal food while the cats are asleep.}\label{fig:rats_steal_before}
	\end{subfigure}
	\quad
	\begin{subfigure}[t]{1.5in}
		\fbox{\includegraphics[width=1.67in]{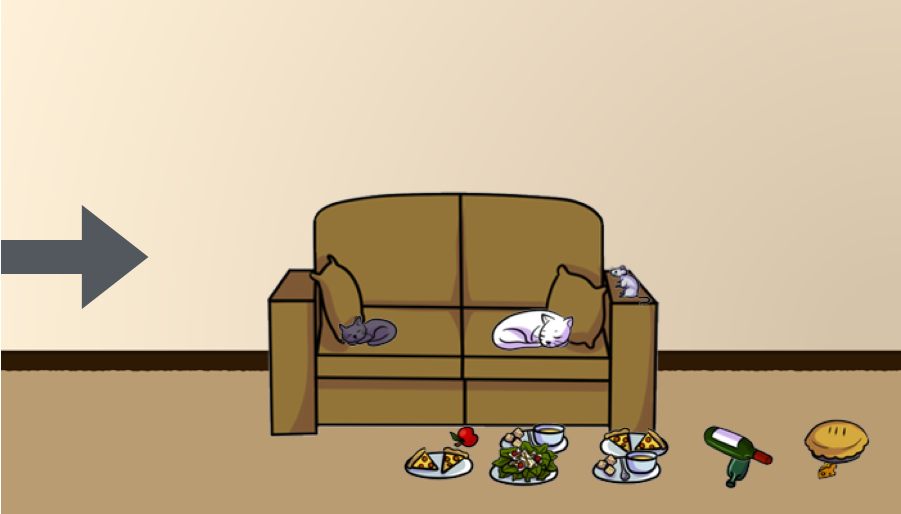}}
		\vspace{-17pt}
		\caption{\emph{Funny Object Replaced (unfunny) counterpart:} Rats in \emph{(c)} are replaced by food.}\label{fig:rats_steal_after}
	\end{subfigure}
\vspace{-0.015\textwidth}
\caption{
(a), (b) are selected funny scenes in the Abstract Visual Humor dataset. (c) is an originally funny scene in the Funny Object Replaced dataset. The objects contributing to humor in (c) are replaced by a human with other objects, to create an unfunny counterpart.
}
\label{fig:teaser}
\vspace{-0.04\textwidth}
\end{figure}
Computational visual humor is useful for a number of applications: to create better photo editing tools, smart cameras that pick the right moment to take a (funny) picture, recommendation tools that rate funny pictures higher (say, to post on social media), video summarization tools that summarize only the funny frames, automatically generating funny scenes for entertainment, identifying and catering to personalized humor, \etc.
As AI systems interact more with humans, it is vital that they understand subtleties of human emotions and expressions. In that sense, being able to identify humor can contribute to their \emph{common sense}. \par
Understanding visual humor is fraught with challenges such as having to detect all objects in the scene, observing the interactions between objects, and understanding context, which are currently unsolved problems.
In this work, we argue that, by using scenes made from clipart~\cite{vqa, interact, ObjectDynamicsScenes, vis-w2v, imagination, commonSense, SemanticsIntoFocus, VisualInterpretationSents}, we can study visual humor without having to wait for these detailed recognition problems to be solved. Abstract scenes are inherently densely annotated (\eg all objects and their locations are known), and so enable us to learn fine-grained semantics of a scene that causes it to be funny. 
In this paper, we collect two datasets of abstract scenes that facilitate the study of humor at both the scene-level (\figref{fig:raccoon_party}, \figref{fig:dog_dinner}) and the object-level (\figref{fig:rats_steal_before}, \figref{fig:rats_steal_after}). We propose a model that predicts how funny a scene is using semantic visual features of the scene such as occurrence of objects, and their relative locations. 
We also build computational models for a particular source of humor, \ie, humor due to the presence of objects in an unusual context. This source of humor is explained by the \emph{incongruity theory} of humor which states that a playful violation of the subjective expectations of a perceiver causes humor \cite{MultidiscplinaryFacetsHumorResearch}. \Eg, \figref{fig:dog_dinner} is funny because our expectation is that people eat at tables and dogs sit in pet beds and this is violated when we see the roles of people and dogs swapped. \par
The scene-level Abstract Visual Humor (AVH) dataset contains funny scenes (\figref{fig:raccoon_party}, \figref{fig:dog_dinner}) and unfunny scenes with human ratings for funniness of each scene. Using the ground truth rating, we demonstrate that we can reliably predict a \emph{funniness score} for a given scene. 
The object-level Funny Object Replaced (FOR) dataset contains scenes that are originally funny (\figref{fig:rats_steal_before}) and their unfunny counterparts (\figref{fig:rats_steal_after}). The unfunny counterparts are created by humans by replacing objects that contribute to humor such that the scene is not funny anymore. The ground truth of replaced objects is used to train models to alter the funniness of a scene -- to make a funny scene unfunny and vice versa. Our models outperform natural baselines and ablated versions of our system in quantitative evaluation. They also demonstrate good qualitative performance via human studies. \par
Our main contributions are as follows:
\begin{compactenum}[1.]
	\item We collect two abstract scene datasets consisting of scenes created by humans which are publicly available.
		\begin{compactenum}[i.]
			\item The scene-level Abstract Visual Humor (AVH) dataset consists of funny and unfunny abstract scenes (\secref{subsec:AVH}). Each scene also contains a brief explanation of the humor in the scene. 
			\item The object-level Funny Object Replaced (FOR) dataset consists of funny scenes and their corresponding unfunny counterparts resulting from object replacement (\secref{subsec:FOR}). 
		\end{compactenum}
	\item We analyze the different sources of humor techniques depicted in the AVH dataset via human studies (\secref{subsec:humor_techniques}). 
	\item We learn distributed representations for each object category which encode the context in which an object naturally appears, \ie, in an unfunny setting. (\secref{subsec:features}). 
	\item We model two tasks to demonstrate an understanding of visual humor:
	\begin{compactenum}[i.]
		\item Predicting how funny a given scene is (\secref{subsec:task1_results}).
		\item Automatically altering the funniness of a given scene (\secref{subsec:task2_funny_unfunny}).
	\end{compactenum}	
\end{compactenum}  \par
To the best of our knowledge, this is the first work that deals with understanding and building computational models for visual humor. 
\vspace{-0.0075\textwidth} 
\section{Related Work}
\label{sec:related_work}
%
\noindent 
\textbf{Humor Theories.}
Humor has been a topic of study since the time of Plato \cite{CollectedDialoguesPlato}, Aristotle \cite{BasicWorksAristotle} and Bharata \cite{NatyaSastra}. Over the years, philosophical studies and psychological research have sought to explain why we laugh. There are three theories of humor \cite{wiki_theoriesOfHumor} that are popular in contemporary academic literature. According to the incongruity theory, a perceiver encounters an incongruity when expectations about the stimulus are violated \cite{incong_vs_incogRes}. The two stage model of humor \cite{TwoStageModelHumor} further states that the process of discarding prior assumptions and reinterpreting the incongruity in a new context (resolution) is crucial to the comprehension of humor.
Superiority theory suggests that the misfortunes of others which reflects our own superiority is a source of humor \cite{HumorResearchStateOfArt}.
According to the relief theory, humor is the release of pent-up tension or mental energy. Feelings of hostility, aggression, or sexuality that are expressed bypassing any societal norms are said to be enjoyed \cite{JokeRelationUnconscious}. \par
Previous attempts to characterize the stimuli that induce humor have mostly dealt with linguistic or verbal humor \cite{MultidiscplinaryFacetsHumorResearch} \eg, script-based semantic theory of humor \cite{SSTH} and its revised version, the general theory of verbal humor \cite{GVTH}.\par
\noindent 
\textbf{Computational Models of Humor.}
A number of computational models are developed to recognize language-based humor~\eg, one-liners~\cite{MakingComputersLaugh}, sarcasm~\cite{sarcasm} and \emph{knock-knock} jokes~\cite{ComputationalWordplayJokes}. Other work in this area includes exploring features of humorous texts that help detection of humor~\cite{CharacterizingHumour}, and identifying the set of words or phrases in a sentence that could contribute to humor~\cite{yanghumor}. \par
Some computational humor models that generate verbal humor are JAPE \cite{JAPE} which is a pun-based riddle generating program, HAHAcronym \cite{HAHAcronym} which is an automatic funny acronym generator, and an unsupervised model that produces \quotes{\textit{I like my X like I like my Y, Z}} jokes \cite{UnsupervisedJokeBigData}.
While the above works investigate detection and generation of verbal humor, in this work we deal purely with \emph{visual} humor.  \par
Recent works 
predict the best text to go along with a given (presumably funny) raw image such as a meme~\cite{meme} or a cartoon~\cite{IdentifyingHumorousCartoonCaptions}. In addition, Radev~\etal~\cite{UnsupervisedCartoonCaption} develop unsupervised methods to rank funniness of captions for a cartoon. They also analyze the characteristics of the funniest captions.
Unlike our work, these works do not predict whether a \emph{scene} is funny or which components of the scene contribute to the humor. \par
Buijzen and Valkenburg \cite{audioVisualTypology} analyze humorous commercials to develop and investigate a typology of humor. Our contributions are different as we study the sources of humor in static images, as opposed to audiovisual media.
To the best of our knowledge, ours is the first work to study \emph{visual} humor in a computational framework. \par
\noindent 
\textbf{Human Perception of Images.}
A number of works investigate the intrinsic characteristics of an image that influence human perception \eg, memorability~\cite{memorability}, popularity~\cite{popularity}, visual interestingness~\cite{interestingness}, and virality~\cite{virality}.
In this work, we study what content in a scene causes people to perceive it as funny, and explore a method of altering the funniness of a scene. \par
\noindent 
\textbf{Learning from Visual Abstraction.}
Visual abstractions have been used to explore high-level semantic scene understanding tasks like identifying visual features that are semantically important~\cite{SemanticsIntoFocus, AdoptingAbstractSemanticUnderstandingPAMI}, learning mappings between visual features and text~\cite{VisualInterpretationSents}, learning visually grounded word embeddings \cite{vis-w2v}, modeling fine-grained interactions between pairs of people~\cite{interact}, and learning (temporal and static) common sense~\cite{ObjectDynamicsScenes, imagination, commonSense}. In this work, we use abstract scenes to understand the semantics in a scene that cause humor, a problem that has not been studied before. 

\section{Datasets}
\label{sec:datasets}
We introduce two new abstract scenes datasets -- the Abstract Visual Humor (AVH) dataset (\secref{subsec:AVH}) and the Funny Object Replaced (FOR) dataset (\secref{subsec:FOR}) using the interfaces described in \secref{subsec:interface}. The AVH dataset (\secref{subsec:AVH}) consists of both funny and unfunny scenes along with funniness ratings. The FOR dataset (\secref{subsec:FOR}) consists of funny scenes and their altered unfunny counterparts. Both the datasets are made publicly available on the project webpage. 
\subsection{Abstract Scenes Interface}
\label{subsec:interface}
Abstract scenes enable researchers to explore high-level semantics of a scene without waiting for low-level recognition tasks to be solved. 
We use the clipart interface\footnote{\url{www.github.com/VT-vision-lab/abstract_scenes_v002}} developed by Antol \etal~\cite{vqa} which allows for indoor and outdoor scenes to be created. The clipart vocabulary consists of 20 deformable human models, 31 animals in various poses, and around 100 objects that are found in indoor (\eg, chair, table, sofa, fireplace, notebook, painting) and outdoor (\eg, sun, cloud, tree, grill, campfire, slide) scenes.
The human models span different genders, races, and ages with 8 different expressions. They have limbs that are adjustable to allow for continuous pose variations. This combined with the large vocabulary of objects result in diverse scenes with rich semantics. \figref{fig:teaser} (\emph{Top Row}) shows scenes that AMT workers created using this abstract scenes interface and vocabulary. Additional details, example scenes, and a sample of clipart objects are available on the project webpage. \par 
\begin{figure*}[t]
\setlength{\fboxsep}{0pt}%
\setlength{\fboxrule}{0pt}%
\begin{center}
	\begin{subfigure}[t]{1.5in}
		\centering
		\fbox{\includegraphics[width=1.65in]{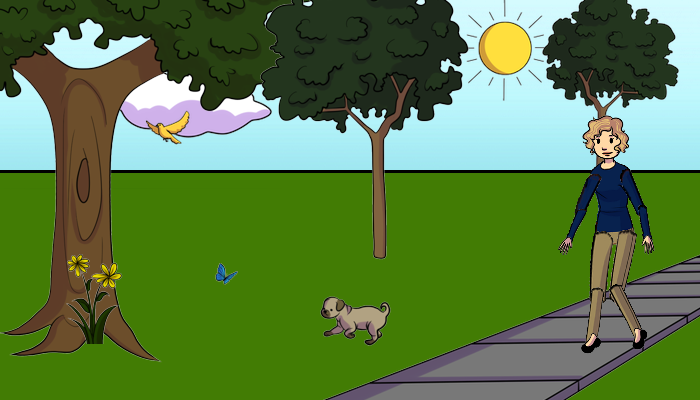}}
		\caption{0.1}\label{fig:boring}
	\end{subfigure}
	\begin{subfigure}[t]{1.5in}
		\centering
		\fbox{\includegraphics[width=1.65in]{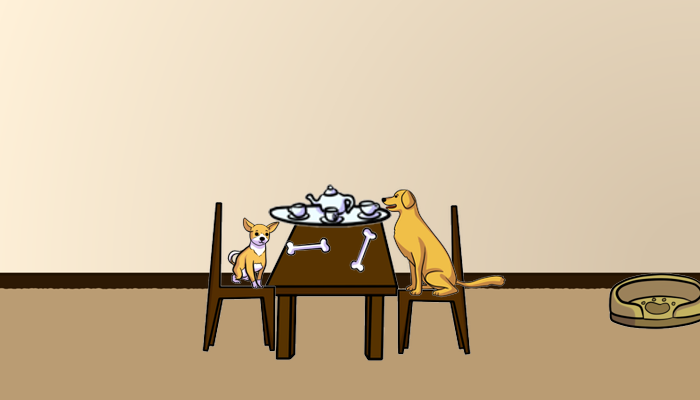}}
		\caption{1.5}\label{fig:boring}
	\end{subfigure}
	\begin{subfigure}[t]{1.5in}
		\centering
		\fbox{\includegraphics[width=1.65in]{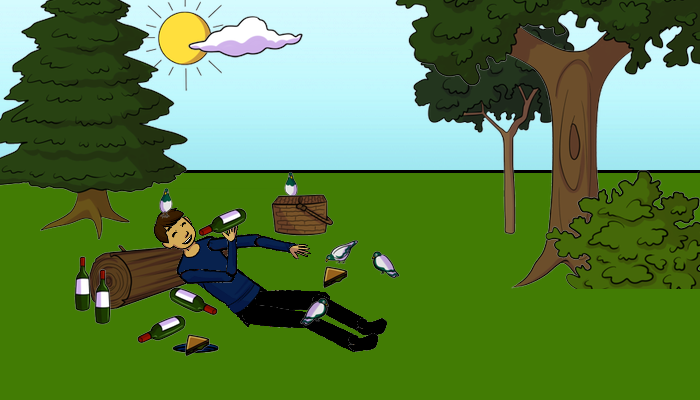}}
		\caption{4.0}\label{fig:boring}
	\end{subfigure}
	\begin{subfigure}[t]{1.5in}
		\centering
		\fbox{\includegraphics[width=1.65in]{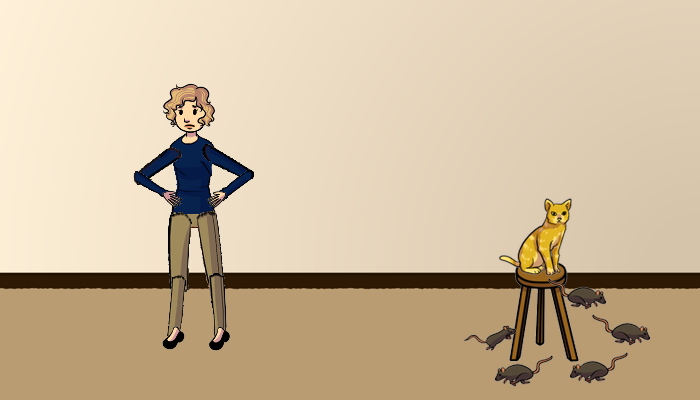}}
		\caption{4.0}\label{fig:boring}
	\end{subfigure}
\end{center}
\vspace{-20pt}
\caption{Spectrum of scenes \emph{(left to right)} in ascending order of funniness score, \funniness\ (\secref{subsubsec:funniness_score}) as rated by AMT workers.}
\label{fig:spectrum_scenes}
\vspace{-15pt}
\end{figure*}
\subsection{Abstract Visual Humor (AVH) Dataset}
\label{subsec:AVH}
This dataset consists of funny and unfunny scenes created by AMT workers, facilitating the study of visual humor at the scene level. \par
\noindent
\textbf{Collecting Funny Scenes. }
We collect 3.2K scenes via AMT by asking workers to create funny scenes that are meaningful, realistic, and that other people would also consider funny. This is to encourage workers to refrain from creating scenes with inside jokes or catering to a very personalized form of humor. A screenshot of the interface used to collect the data is available on the project webpage.
We provide a random subset of the clipart vocabulary to each worker out of which at least 6 clipart objects are to be used to create a scene. 
In addition, we also ask the worker to give a brief description of why the scene is funny in a short phrase or sentence. We find that this encourages workers to be more thoughtful and detailed regarding the scene they create. Note that this is different from providing a caption to an image since this is a simple explanation of what the worker had in mind while creating the scene. Mining this data may be useful to better understand visual humor. However, in this work we focus on the harder task of understanding purely \emph{visual} humor and do not use these explanations. \par
We also use an equal number (3.2K) of abstract scenes from \cite{vqa} which are realistic, everyday scenes. We expect most of these scenes to be mundane (\ie., not funny). \par
\noindent
\textbf{Labeling Scene Funniness. }
\label{subsubsec:funniness_score}
Anyone who has tried to be funny knows that humor is a subjective notion. A well-intending worker may create a scene that other people do not find very funny. We obtain funniness ratings for each scene in the dataset from 10 different workers on AMT who do not see the creator's explanation of funniness. The ratings are on a scale of 1 to 5, where 1 is not funny and 5 is extremely funny. 
We define the \emph{funniness score} \funniness\, of a scene \textit{i}, as the average of the 10 ratings for the scene. We found 10 ratings to be sufficient for good inter-human agreement. Further analysis is provided on the project webpage. \par
By plotting a distribution of these scores, we determine the optimal threshold that best separates scenes that were intended to be funny (\ie, workers were specifically asked to create a funny scene) and other scenes (\ie, everyday scenes from~\cite{vqa}, where workers were not asked to create funny scenes). 
We label all scenes that have a $\funniness\geqslant \text{threshold}$ as \emph{funny} and all scenes with a lower \funniness\ as \emph{unfunny}. 
This re-labeling results in 522 \emph{unintentionally funny} scenes (i.e., scenes from~\cite{vqa}, which were determined to be funny), and 682 \emph{unintentionally unfunny} scenes (\ie, well-intentioned worker outputs which were deemed not funny by the crowd). \par
In total, this dataset contains 6,400 scenes (3,028 funny scenes and 3,372 unfunny scenes). We randomly split these scenes into train, val, and test sets having 60\%, 20\%, and 20\% of the scenes, respectively. We refer to this dataset as the AVH dataset. \par
\noindent
\textbf{Humor Techniques. }
\label{subsec:humor_techniques}
To better understand the different sources of humor in our dataset, we collect human annotations of the different techniques are used to depict humor in each scene. We create a list of humor techniques that are motivated by existing humor theories, based on patterns that we observe in funny scenes, and the audio-visual humor typology by Buijzen \etal~\cite{audioVisualTypology}: 
\emph{person doing something unusual}, \emph{animal doing something unusual}, \emph{clownish behavior (\ie, goofiness)}, \emph{too many objects}, \emph{somebody getting hurt}, \emph{somebody getting scared} and \emph{somebody getting angry}.  \par
\begin{figure*}[t!]
\centering
\setlength{\fboxsep}{0pt}
\setlength{\fboxrule}{0pt}
\fbox{\includegraphics[width=0.24\textwidth]{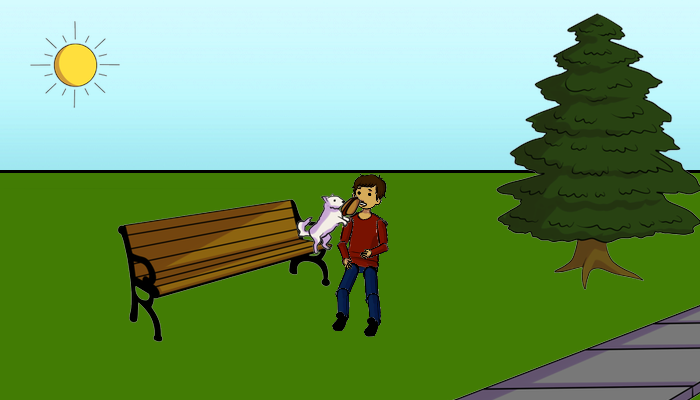}\hspace{0.005\textwidth}}%
\fbox{\includegraphics[width=0.24\textwidth]{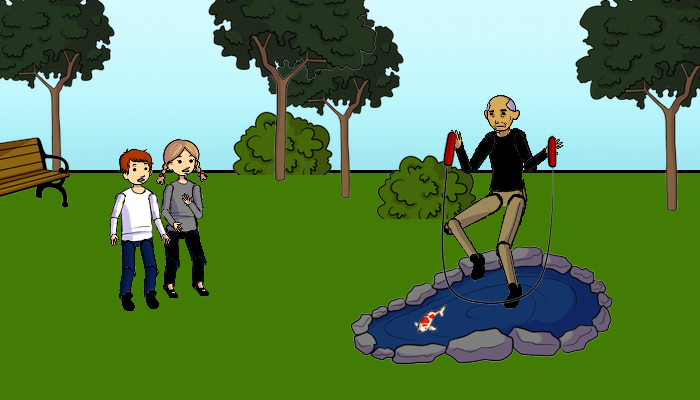}\hspace{0.005\textwidth}}%
\fbox{\includegraphics[width=0.24\textwidth]{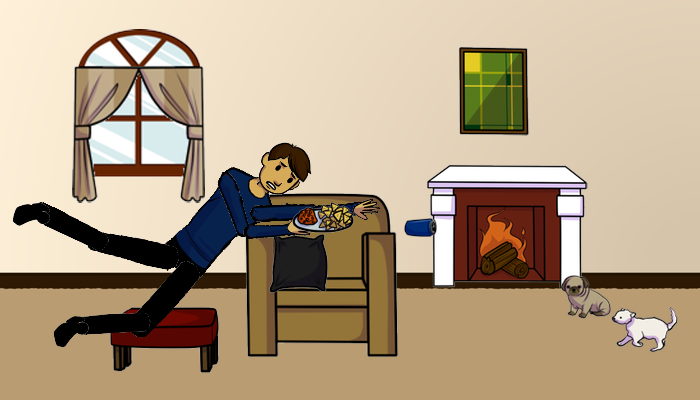}\hspace{0.005\textwidth}}%
\fbox{\includegraphics[width=0.24\textwidth]{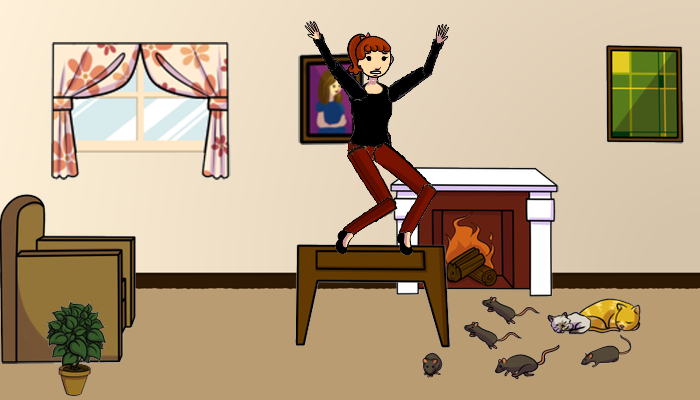}}
\vspace{-3pt}
\caption{Top voted scenes by humor technique (\secref{subsec:humor_techniques}). From \emph{left} to \emph{right}: \emph{animal doing something unusual}, \emph{person doing something unusual}, \emph{somebody getting hurt}, and \emph{somebody getting scared}.}
\label{fig:humor_techniques}
\vspace{-2pt}
\end{figure*}
We choose a subset of 200 funny scenes from the AVH dataset. We show each of these scenes to 10  different AMT workers and ask them to choose all the humor techniques that are depicted. Our options also included \emph{none of the above reasons}, which also prompted workers to briefly explain what other unlisted technique depicted in the scene made it funny. 
However, we observe that this option was rarely used by workers. This may indicate that most of our scenes can be explained well by one of the listed humor techniques.
\figref{fig:humor_techniques} shows the top voted images corresponding to the 4 most popular techniques of humor. We find that the techniques that involve animate objects -- \emph{animal doing something unusual} and \emph{person doing something unusual} are voted higher than any other technique by a large margin. For 75\% of the scenes, at least 3 out of 10 workers picked one of these two techniques. We observe that this \emph{unusualness} or \emph{incongruity} is generally caused by objects occurring in an unusual context in the scene.  \par
Introducing or eliminating incongruities can alter the funniness of a scene. An elderly person kicking a football while simultaneously skateboarding (\figref{fig:before_after}, \emph{bottom}) is incongruous and hence considered funny. However, when the person is replaced by a young girl, this is is not incongruous and hence not funny. Such incongruities that can alter the funniness of a scene serves as our motivation to collect the Funny Object Replaced dataset which we describe next. 
\subsection{Funny Object Replaced (FOR) Dataset}
\label{subsec:FOR}
Replacing objects in a scene is a technique to manipulate incongruities (and hence funniness) in a scene. For instance, we can change funny interactions (which are unexpected by our common sense) to interactions that are \emph{normal} according to our mental model of the world. We use this technique to collect a dataset which consists of funny scenes and their altered unfunny counterparts. This enables the study of humor in a scene at the \emph{object-level}. \par 
We show funny scenes from the AVH dataset and ask AMT workers to make the least number of replacements in the scene to render the originally funny scene unfunny. The motivation behind this is to get a precise signal of which objects in the scene contribute to humor and what they can be replaced with to reduce/eliminate humor, while keeping the underlying structure of the scene the same. We ask workers to replace an object with another object that is as similar as possible to the first object and keep the scene realistic. This helps us understand fine-grained semantics that causes a specific object category to contribute to humor. There could be other ways to manipulate humor, \eg, by adding, removing, or moving objects in a scene, \etc but in our work we employ only the technique of replacing objects. We find that this technique is very effective in altering the funniness of a scene. Our interface did not allow people to add, remove, or move the objects in the scene. A screenshot of the interface used to collect this dataset is available on the project webpage. \par
\begin{figure}[t!]
\centering
\setlength{\fboxsep}{0pt}%
\setlength{\fboxrule}{0pt}%
\fbox{\includegraphics[width=0.24\textwidth]{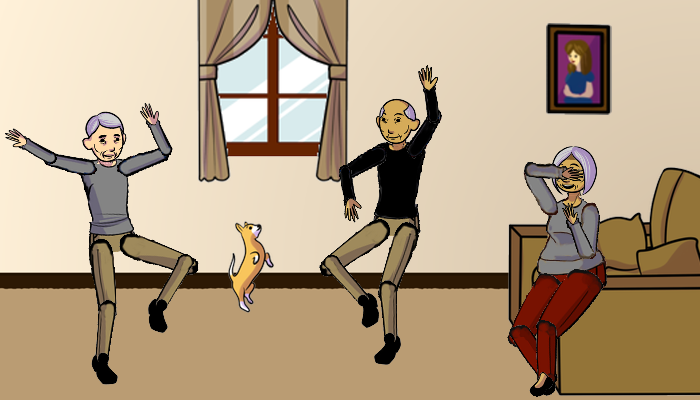}\hspace{0.005\textwidth}}%
\fbox{\includegraphics[width=0.24\textwidth]{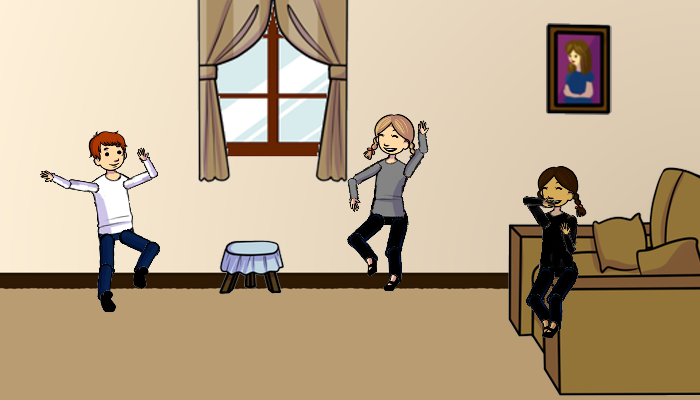}}\vspace{0.004\textwidth}
\fbox{\includegraphics[width=0.24\textwidth]{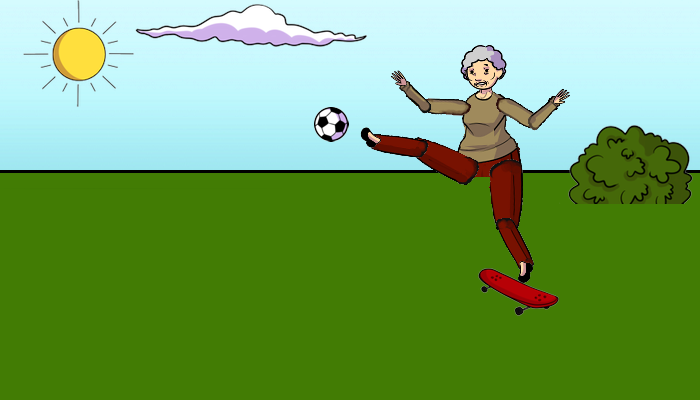}\hspace{0.005\textwidth}}%
\fbox{\includegraphics[width=0.24\textwidth]{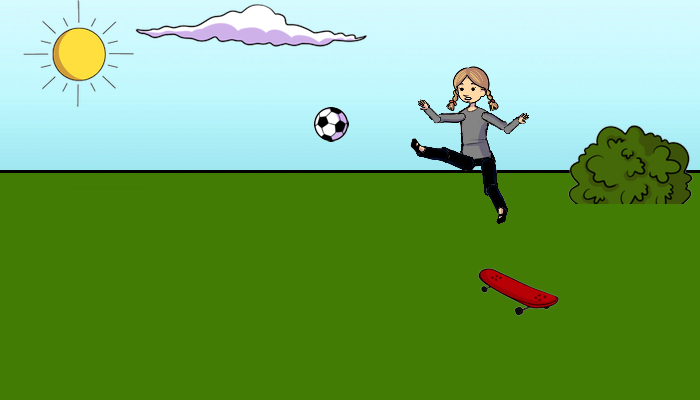}}%
\caption{Funny scenes (\emph{left}) and \emph{one} among the 5 corresponding object-replaced unfunny counterparts (\emph{right}) from the FOR dataset (see \secref{subsec:FOR}). For each funny scene, we collect an unfunny counterpart from a different worker.}
\label{fig:before_after}
\vspace{-12pt}
\end{figure}

For each of the 3,028 funny scenes in the AVH dataset, we collect \emph{object-replaced} scenes from 5 different workers resulting in 15,140 unfunny counterpart scenes. As a sanity check, we collect funniness ratings (via AMT) for 750 unfunny counterpart scenes. We observe that they indeed have an average \funniness\ of 1.10, which is smaller than that of their corresponding original funny scenes (whose average \funniness\ is 2.66). \figref{fig:before_after} shows two pairs of funny scenes and their object-replaced unfunny counterparts. We refer to this dataset as the FOR dataset. \par
Given the task posed to workers (altering a funny scene to make it unfunny), it is natural to use this dataset to train a model to reduce the humor in a scene. However, this dataset can also be used to train flipped models that can increase the humor in a scene as shown in \secref{subsubsec:task2_unfunny_funny}.
\section{Approach}
\label{sec:approach}
\vspace{-0.01\textwidth}
We propose and model two tasks that we believe demonstrate an understanding of some aspects of visual humor: 
\begin{compactenum}[1.]
\item Predicting how funny a given scene is. 
\item Altering the funniness of a scene. 
\end{compactenum}
The models that perform the above tasks are described in \secref{subsec:task1} and \secref{subsec:task2}, respectively. The features used in the models are described first (\secref{subsec:features}).
\subsection{Features}
\label{subsec:features}
Abstract scenes are trivially densely annotated which we use to compute rich semantic features. Recall that our interface allows two types of scenes (indoor and outdoor) and our vocabulary consists of 150 object categories. We compute both scene-level and instance-level features. \par
\vspace{0.005\textwidth}
\begin{compactenum}[1.]
\item \textbf{Instance-Level Features}
\begin{enumerate}[leftmargin=-0.5cm,itemindent=0cm,labelwidth=\itemindent,labelsep=0cm,align=left]
\item \textbf{ Object embedding (150-d)}
\label{subsec:objEmbedding}
 is a distributed representation that captures the context in which an object category usually occurs. We learn this representation using a word2vec-style continuous Bag-of-Words model \cite{word2vec}. 
The model tries to predict the presence of an object category in the scene, given the context provided by other instances of objects in the scene. Specifically, in a scene, given 5 (randomly chosen) instances, the model tries to predict the object category of the 6th instance. 
We train the single-layer (150-d) neural network \cite{word2vec_first} with multiple 6-item subsets of instances from each scene.
The network is trained using Stochastic Gradient Descent (SGD) with a momentum of 0.9. We use 11K scenes (that were not intended to be funny) from the dataset collected in~\cite{vqa} to train the model. Thus, we learn representations of objects occurring in natural contexts which are not funny.
A visualization of the object embeddings is available on the project webpage. 
\item \textbf{ Local embedding  (150-d)} For each instantiation of an object in the scene, we compute a weighted sum of object embeddings of all the other instances in the scene. The weight of every other instance is its inverse square-root distance \wrt the instance under consideration. \newline
\end{enumerate} 
\item \textbf{Scene-Level Features} 
	\begin{enumerate}[leftmargin=-0.5cm,itemindent=0cm,labelwidth=\itemindent,labelsep=0cm,align=left]
	\item \textbf{ Cardinality (150-d)} is a Bag-of-Words representation that indicates the number of instances of each object category that are present in the scene.
	\item \textbf{ Location (300-d)} is a vector of the horizontal and vertical coordinates of every object in the scene. 
	When multiple instances of an object category are present, we consider location of the instance closest to the center of the scene.
	\item \label{sc_emb} \textbf{ Scene Embedding (150-d)} is the sum of object embeddings of all objects present in the scene.
	\end{enumerate}
\end{compactenum}
\subsection{Predicting Funniness Score}
\label{subsec:task1}
We train a Support Vector Regressor (SVR) that predicts the funniness score, \funniness\ for a given scene $i$. The model regresses to the \funniness\ computed from ratings given by AMT workers (described in \secref{subsubsec:funniness_score}) on scenes from the AVH dataset (\secref{subsec:AVH}). 
We train the SVR on the scene-level features (described in \secref{subsec:features}) and perform an ablation study. 
\subsection{Altering Funniness of a Scene}
\label{subsec:task2}
We learn models to alter the funniness of a scene -- from funny to unfunny and \emph{vice versa}. Our two-stage pipeline 
involves:
\begin{compactenum}[1.]
\item Detecting objects that contribute to humor. 
\item Identifying suitable replacement objects from 1. to make the scene unfunny (or funny), while keeping it realistic.
\end{compactenum}
\noindent
\textbf{Detecting Humor. }We train a multi-layer perceptron (MLP) on scenes from the FOR dataset to make a binary prediction on each object instance in the scene -- whether it should be replaced to alter the funniness of a scene or not. The input is a 300-d vector formed by concatenating object embedding and local embedding features. The MLP has two hidden layers comprising of 300 and 100 units respectively, to which ReLU activation is applied. The final layer has 2 neurons and is used to perform binary classification (replace or not) using cross-entropy loss.
We train the model using SGD with a base learning rate of 0.01 and momentum of 0.9. We also trained a model with skip-connections that considers the predictions made on other objects when making a prediction on a given object. However, this did not result in significant performance gains. \par
\noindent
\textbf{Altering Humor. }
We train an MLP 
to perform a 150-way classification to predict potential replacer objects (from the clipart vocabulary), given an object predicted to be replaced in a scene. The model's input is a 300-d vector formed by concatenating local embedding and object embedding features. The classifier has 3 hidden layers of 300 units each, with ReLU non-linearities. The output layer has 150 units over which we compute soft-max loss. We train the model using SGD with a base learning rate of 0.1, momentum of 0.9, and a dropout ratio of 0.5. The label for an instance is the index of the replacer object category used by the worker. Due to the large diversity of viable replacer objects that can alter humor in a scene, we also analyze the top-5 predictions of this model.
We train two models -- one on funny scenes, and another on their unfunny counterparts from the FOR dataset. Thus, we learn models to alter the funniness in a scene in one direction -- funny to unfunny or vice versa. Although we could train the pipeline end-to-end, we train each stage separately so that we can evaluate them separately and isolate their errors (for better interpretability). 
\section{Results}
\label{sec:results}
We discuss the performance of our models in the two visual humor tasks of: 
\begin{compactenum}
	\item Predicting how funny a given scene is (\secref{subsec:task1_results}) 
	\item Altering funniness of a scene (\secref{subsec:task2_funny_unfunny}). 
\end{compactenum}
We discuss the quantitative results of our model in altering an unfunny scene to make it funny in \secref{subsubsec:task2_funny_unfunny}), and the \emph{vice versa} in \secref{subsubsec:task2_unfunny_funny}. In \secref{subsec:human_eval}, we report qualitative results through human studies.
\subsection{Predicting Funniness Score} 
\label{subsec:task1_results}
This section presents performance of the SVR (\secref{subsec:task1}) that predicts the funniness score \funniness\ of a scene. \par
\noindent
\textbf{Metric.} 
We use average relative error to quantify our model's performance computed as follows:
\begin{equation} 
\frac{1}{N}\sum_{i=1}^{N}\frac{|Predicted\ \funniness - Ground\ Truth\ \funniness|}{Ground\ Truth\ \funniness}
\end{equation}
where \emph{N} is the number of test scenes and \funniness\ is the funniness score for the test scene \emph{i}.  \par
\noindent
\textbf{Baseline:} 
The baseline model always predicts the average funniness score of the training scenes.  \par
\noindent
\textbf{Model.} 
As shown in \tableref{tab:reg}, we observe that our model trained using combinations of different scene-level features (described in \secref{subsec:features}) performs better than the baseline model. We see that Location features perform slightly better than Cardinality. This makes sense because Location features also have occurrence information. The Embedding does not have location information and hence does worse. Due to some redundancy (all features have occurrence information), combining them does not improve performance.
\begin{table}[t]
\setlength{\tabcolsep}{3.2pt}
{\small
\begin{center}
\begin{tabular}{@{}llcccc@{}}
\toprule
\textbf{Features} & \textbf{Avg. Rel. Err.}  \\ 
\midrule
Avg. Prediction Baseline &   \multicolumn{1}{r}{0.3151} \\
Embedding & \multicolumn{1}{r}{0.2516} \\
Cardinality & \multicolumn{1}{r}{0.2450} \\ 
Location & \multicolumn{1}{r}{0.2400}  \\ 
Embedding + Cardinality + Location & \multicolumn{1}{r}{0.2400}\\
\bottomrule
\end{tabular}
\end{center}
\vspace{-0.025\textwidth} 
\caption {Performance of different feature combinations in predicting funniness score \funniness\ of a scene.}
\label{tab:reg}
\vspace{-0.02\textwidth} 
}
\end{table}
\vspace{-0.025\textwidth} 
\subsection{Altering Funniness of a Scene} 
\label{subsec:task2_funny_unfunny}
\vspace{-0.0025\textwidth} 
We discuss the performance in the tasks of identifying objects in a scene that contribute to humor (\secref{subsec:task1}) and replacing those objects with other objects to reduce (or increase) humor (\secref{subsec:task2}).
\vspace{-0.020\textwidth} 
\subsubsection{Predicting Objects to be Replaced}
\label{subsec:predicting_objs_replaced}
We train this model to detect objects instances that are funny in the scene. It makes a binary prediction whether each instance should be replaced or not. \newline
\noindent
\textbf{Metric.} Along with \naive accuracy (\% of correct predictions, \ie, Acc.), we also report average class-wise accuracy (\ie, Avg. Cl. Acc.) to determine the performance of our model for this task. As the data is skewed, with the majority class being \emph{not-replace}, we require our model to perform well both class-wise and as a whole. \newline
\noindent
\textbf{Baselines:} 
\begin{compactenum}[1.]
\item \textbf{ Priors.} We always predict that an instance should not be replaced. We also compute a stronger baseline that replaces an object if it is replaced at least T\% of the time in training data. T was set to 20 based on the validation set.
\item \textbf{ Anomaly Detection.} From the scene embedding, we subtract the object embedding of the object under consideration. We then compute the cosine similarity of the resultant scene embedding with the object embedding. Objects with the least similarity with the scene are the anomalous objects in the scene.
This is similar to finding the odd-one-out given a group of words~\cite{word2vec_first}. Objects that have a cosine similarity less than a threshold T with the scene are predicted as anomalous objects and are replaced. A modification to this baseline is to replace K objects that are least similar to the scene. Based on performance on the validation set, T and K are determined to be 0.8 and 4, respectively.
\end{compactenum}
\noindent
\textbf{Model.}
\tableref{tab:model1} compares the performance of our model with the baselines described above. We observe that the baseline based on priors performs better than anomaly detection. This is perhaps not surprising because the prior-based baseline, while \naive, is \quotes{supervised} in the sense that it relies on statistics from the training dataset of which objects tend to get replaced. On the other hand, anomaly detection is completely unsupervised since it only captures the context of objects in \emph{normal} scenes. Our approach performs better than the baseline approaches in identifying objects that contribute to humor. \par
On average, we observe that our model replaces 3.67 objects for a given image as compared to an average of 2.54 objects replaced in the ground truth. This bias to replace more objects ensures that a given scene becomes significantly less funny than the original scene. 
We observe that the model learns that in general, animate objects like humans and animals are potentially stronger sources of humor compared to inanimate objects. It is interesting to note that the model also learns fine-grained detail, \eg, to replace older people playing outdoors (which may be considered funny) with younger people (\figref{fig:funny2unfunnyr}, top row). 
\vspace{-0.01\textwidth} 
\subsubsection{Making a Scene Unfunny}
\label{subsubsec:task2_funny_unfunny}
\vspace{-0.0025\textwidth} 
Given that an object is predicted to be replaced in the scene, the model has to also predict a suitable replacer object. In this section, we discuss the performance of the model in predicting these replacer objects. 
This model is trained and evaluated using ground truth annotations of objects that are replaced by humans in a scene. This helps us isolate performance between predicting \emph{which objects to replace} and predicting \emph{suitable replacers} . \par
\noindent
\textbf{Metric.}
In order to evaluate the performance of the model on the task of replacing funny objects in the scene to make it unfunny, we use the top-5 metric (similar to ImageNet \cite{russakovsky2015imagenet}), \ie, if any of our 5 most confident predictions match the ground truth, we consider that as a correct prediction. \newline
\noindent
\textbf{Baselines:} 
\begin{compactenum}[1.]
\item \textbf{Priors.} Every object is replaced by one of its 5 most frequent replacers in the training set. \par
\item \textbf{Anomaly Detection.} We subtract the embedding of the object that is to be replaced from the scene embedding. The 5 objects from the clipart vocabulary that are most similar (in the embedding space) to this resultant scene embedding are the ones that contextually \quotes{fit in}. 
\end{compactenum}
\begin{table}[t]
\setlength{\tabcolsep}{3.2pt}
{\small
\begin{center}
\begin{tabular}{@{}llcccc@{}}
\toprule
\textbf{Method} & \textbf{Avg. Cl. Acc.} & \textbf{Acc.}   \\
\midrule
Priors (do not replace) & \multicolumn{1}{r}{50\%}\% & \textbf{79.86}\%  \\ 
Priors (object's tendency to be replaced) & \multicolumn{1}{r}{73.13}\%  &  71.5\%\\
\midrule
Anomaly detection (threshold distance) & \multicolumn{1}{r}{62.16}\%  & 58.30\% \\
Anomaly detection (top-K objects) & \multicolumn{1}{r}{63.01}\% & 64.31\% \\
\midrule
Our model & \multicolumn{1}{r}{\textbf{74.45\%}} & 74.74\%  \\
\bottomrule
\end{tabular}
\end{center}
}
\vspace{-0.03\textwidth} 
\caption {Performance of predicting whether an object should be replaced or not, for the task of altering a funny scene to make it unfunny. As the data is skewed with the majority class being \quotes{not-replace}, we require our model to perform well both class-wise and as a whole.}
\label{tab:model1}
\vspace{-0.02\textwidth} \end{table}
\noindent
\textbf{Model.}
We observe that the performance trend in \tableref{tab:model2} is similar to that observed in the previous section (\secref{subsec:predicting_objs_replaced}), \ie., our model performs better than priors, which performs better than anomaly detection. By qualitative inspection, we find that our top prediction is intelligent, but lazy. It eliminates humor in most scenes by choosing to replace objects contributing to humor with other objects that blend well into the background. By relegating an object to the background, it is rendered inactive and hence, cannot be contribute to humor in the scene. For \eg, the top prediction is frequently \quotes{plant} in indoor scenes and \quotes{butterfly} in outdoor scenes. The 2nd prediction is both intelligent and creative. It effectively reduces humor while also ensuring diversity of replacer objects.
Subsequent predictions from the model tend to be less meaningful. Qualitatively, we find the 2nd most confident prediction to be the best compromise. \par
\begin{figure}[t!]
\centering
\setlength{\fboxsep}{0pt}%
\setlength{\fboxrule}{0pt}%
\fbox{\includegraphics[width=0.24\textwidth]{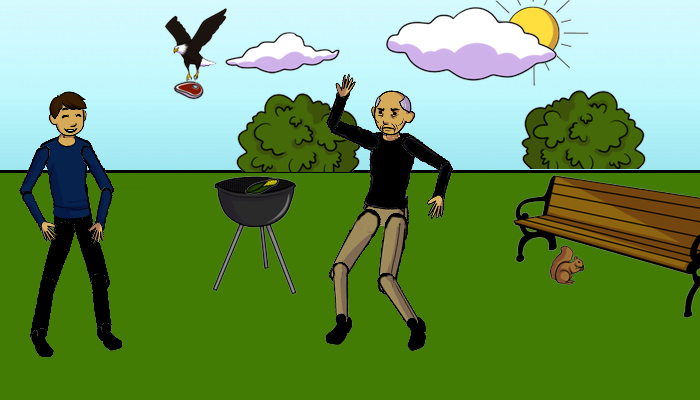}\hspace{0.005\textwidth}}%
\fbox{\includegraphics[width=0.24\textwidth]{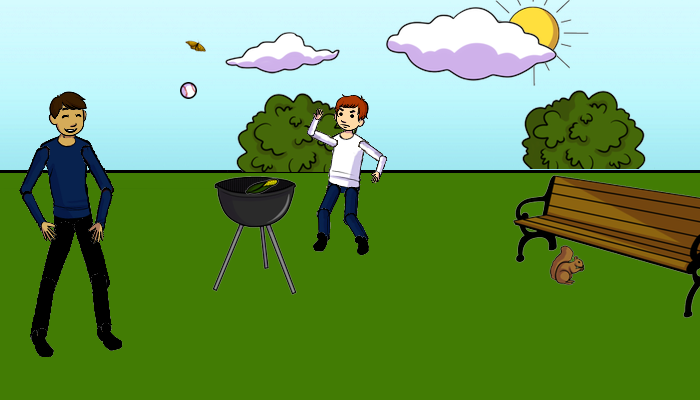}}\vspace{0.004\textwidth}
\fbox{\includegraphics[width=0.24\textwidth]{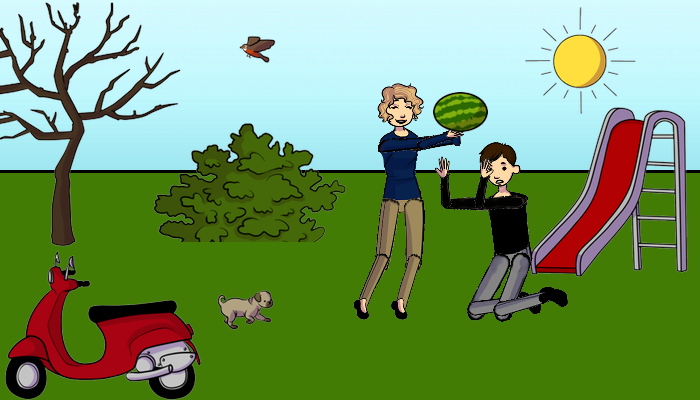}\hspace{0.005\textwidth}}%
\fbox{\includegraphics[width=0.24\textwidth]{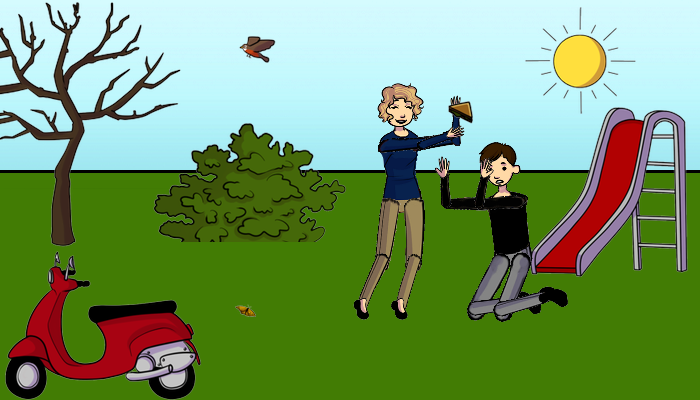}}
\caption{Fully automatic result of altering an input funny scene \emph{(left)} into an unfunny scene \emph{(right)}.}  
\label{fig:funny2unfunnyr}
\vspace{-0.02\textwidth}
\end{figure}

\begin{table}[t]
\setlength{\tabcolsep}{3.2pt}
{\small
\begin{center}
\begin{tabular}{@{}llcccc@{}}
\toprule
\textbf{Method} & \textbf{Top-5 accuracy}  \\
\midrule
Priors (top 5 GT replacers) & \multicolumn{1}{r}{24.53\%}  \\
Anomaly detection (object that \quotes{fits} into scene) & \multicolumn{1}{r}{7.69\%}  \\ 
Our model & \multicolumn{1}{r}{\textbf{29.65\%}}  \\
\bottomrule
\end{tabular}
\end{center}
}
\vspace{-0.03\textwidth} 
\caption {Performance of predicting which object to replace with, for the task of altering a funny scene to make it unfunny.}
\label{tab:model2}
\end{table}
\noindent
\textbf{Full pipeline.} 
\figref{fig:funny2unfunnyr} shows qualitative results from our full pipeline (predicting objects to replace and predicting their replacers) using the 2nd predictions made by our model. 
\vspace{-0.0125\textwidth} 
\subsubsection{Making a Scene Funny}
\label{subsubsec:task2_unfunny_funny}
\begin{figure}[t!]
\centering
\setlength{\fboxsep}{0pt}%
\setlength{\fboxrule}{0pt}%
\fbox{\includegraphics[width=0.24\textwidth]{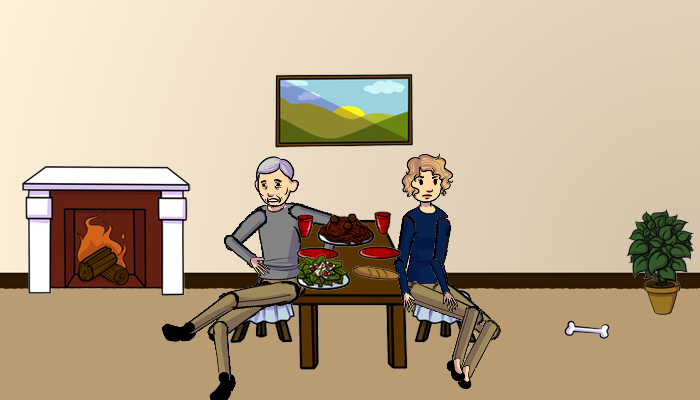}\hspace{0.005\textwidth}}%
\fbox{\includegraphics[width=0.24\textwidth]{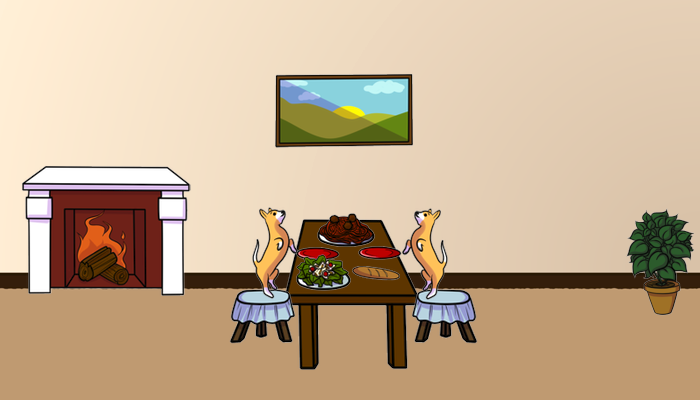}}
\fbox{\includegraphics[width=0.24\textwidth]{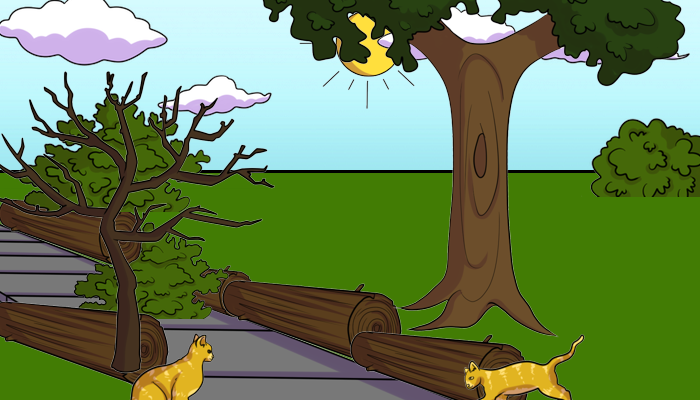}\hspace{0.005\textwidth}}%
\fbox{\includegraphics[width=0.24\textwidth]{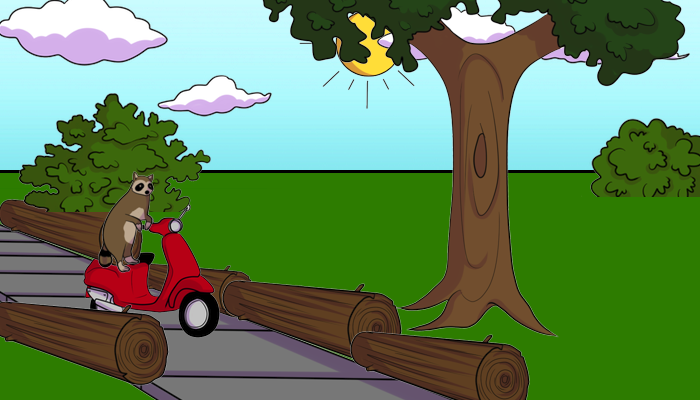}}\vspace{0.004\textwidth}
\caption{Fully automatic result of altering an input unfunny scene \emph{(left)} into a funny scene \emph{(right)}.}  
\label{fig:unfunny2funny}
\vspace{-0.02\textwidth} 
\end{figure}

\vspace{-0.0025\textwidth} 
We train our full pipeline model used in \secref{subsubsec:task2_funny_unfunny} on scenes from the FOR dataset to perform the task of altering an unfunny scene to make it funny. Some qualitative results are shown in \figref{fig:unfunny2funny}.
\subsection{Human Evaluation}
\label{subsec:human_eval}
We conducted two human studies to evaluate our full pipeline: 
\begin{compactenum}[1.]
\item \textbf{Absolute:} We ask 10 workers to rate the funniness of the scene predicted by our model on a scale of 1-5. We then compare this with the \funniness\ of the input funny scene. 
\item \textbf{Relative:} We show 5 workers the input scene and the predicted scene (in random order) and ask them to indicate which scene is funnier. 
\end{compactenum}
\noindent
\textbf{Funny to unfunny.} 
As expected, the output scenes from our model are less funny than the input funny scenes on average. The average \funniness\ of the input funny test scenes is 2.69. This is 1.05 points higher than the output unfunny scenes whose average \funniness\ is 1.64. Unsurprisingly, in relative evaluation, workers find our output scenes to be less funny than the input funny scenes 95\% of the time. \newline
\noindent
\textbf{Unfunny to funny.} During absolute evaluation, we find that the average \funniness\ of scenes made funny by our model is 2.14. This is a relatively high score, considering that the average \funniness\ score of the corresponding originally funny scenes that were created by workers is 2.69. Interestingly, the relative evaluation can be perceived as a \emph{Turing test} of sorts, where we show workers the model's output funny scene and the original funny scene created by workers. 28\% of the time, workers picked the model's scenes to be \emph{funnier}. 
\vspace{-0.0075\textwidth} 
\section{Discussion}
\vspace{-0.005\textwidth} 
Humor is a subtle and complex human behavior. It has many forms ranging from slapstick which has a simple physical nature, to satire which is nuanced and requires an understanding of social context \cite{wikipedia}. Understanding the entire spectrum of humor is a challenging task. It demands perception of fine-grained differences between seemingly similar scenarios. \Eg, a teenager falling off his skateboard (such as in America's Funniest Home Videos\footnote{\url{www.afv.com}}) could be considered funny but an old person falling down the stairs is typically horrifying. Due to these challenges some people even consider computational humor to be an \quotes{AI-complete} problem \cite{computationalHumor, humorReverseEngineerMind}.  \par
While understanding fine-grained semantics is important, it is interesting to note that there exists a qualitative difference in the way humor is perceived in abstract and real scenes. Since abstract scenes are not photorealistic, they afford us \quotes{suspension of reality}. Unlike real images, the content depicted in an abstract scene is benign. Thus, people are likely to find the depiction more funny \cite{BVT}. In our everyday lives, we come across a significant amount of humorous content in the form of comics and cartoons to which our computational models of humor are directly applicable. They can also be applied to learn semantics that can extend to photorealistic images as demonstrated by Antol~\etal~\cite{interact}. \par
Recognizing funniness involves violation of our mental model of how the world \quotes{ought to be} \cite{MultidiscplinaryFacetsHumorResearch}. In verbal humor, the first few lines of the joke (set-up) build up the world model and the last line (punch line) goes against it. It is unclear what forms our mental model when we look at images. Is it our priors about the world around us formed from our past experiences? Is it because we attend to different regions of the image when we look at it and gradually build an expectation of what to see in the rest of the image? These are some interesting questions regarding visual humor that remain unanswered. \par
\vspace{-0.0075\textwidth} 
\section{Conclusion}
\vspace{-0.005\textwidth} 
In this work, we take a step towards understanding and predicting visual humor. We collect two datasets of abstract scenes which enable the study of humor at different levels of granularity. We train a model to predict the \emph{funniness score} of a given scene. We also explore the different sources of humor depicted in the funny scenes via human studies. We train models using incongruity-based humor to alter a scene's funniness. The models learn that in general, animate objects like humans and animals contribute more to humor compared to inanimate objects. Our model outperforms a strong anomaly detection baseline, demonstrating that detecting humor involves something more than just anomaly detection. In human studies of the task of making an originally funny scene unfunny, humans find our model's output to be less funny 95\% of the time. In the task of making a normal scene funny, our evaluation can be interpreted as a \emph{Turing test} of sorts. Scenes made funny by our model were found to be funnier 28\% of the time when compared with the original funny scenes created by workers. Note that our model would match humans at 50\%. We hope that addressing the problem of studying visual humor using abstract scenes and the two datasets that are made public would stimulate further research in this new direction. \par %
\vspace{0.0075\textwidth} 
\noindent
\textbf{Acknowledgements. }We thank the anonymous reviewers for their valuable comments and suggestions. This work was supported in part by the Paul G. Allen Family Foundation via an award to D.P. DB was partially supported by the National Science Foundation CAREER award, the Army Research Office YIP award, and an Office of Naval Research grant N00014-14-1-0679. The views and conclusions contained herein are those of the authors and should not be interpreted as necessarily representing the official policies or endorsements, either expressed or implied, of the U.S. Government or any sponsor. We thank Xinlei Chen for his work on earlier versions of the clipart interface.  \par

\vspace{15pt}
\noindent
{\fontsize{12}{12}\selectfont \textbf{Overview Of Appendix}} \par
\vspace{5pt}
In the following appendix we provide:
\begin{compactenum}[I.]
\item Inter-human agreement on funniness ratings in the Abstract Visual Humor (AVH) dataset. 
\item Details of the model architecture used to learn object embeddings and visualizations of its embeddings. %
\item A sample of objects from the abstract scenes vocabulary. 
\item Examples of scenes from our datasets.
\item Analysis of occurrences of different object types in scenes from our datasets. 
\item The user interfaces used to collect scenes for the AVH and Funny Object Replaced (FOR) datasets. \newline \par 
\end{compactenum}  
\noindent
{\fontsize{12}{12}\selectfont \textbf{Appendix I: Inter-human Agreement}} \par
\label{subsec:interhumanAg}
\vspace{5pt}
In this section, we describe our experiment to determine inter-human agreement in funniness ratings of scenes. The Abstract Visual Humor (AVH) dataset contains 3,028 funny scenes and 3,372 unfunny scenes that were created by Amazon Mechanical Turk (AMT) workers. The funniness of each scene in the dataset is rated by 10 different workers on a scale of 1-5. We define the \emph{funniness score} of a scene, as the average of all ratings for a scene. In this section, we investigate the extent to which people agree regarding the funniness of a scene.  \par
\begin{figure}
\center
\setlength{\fboxsep}{0pt}%
\setlength{\fboxrule}{0pt}%
\vspace{-0.01\textwidth}
\fbox{\includegraphics[width=0.45\textwidth]{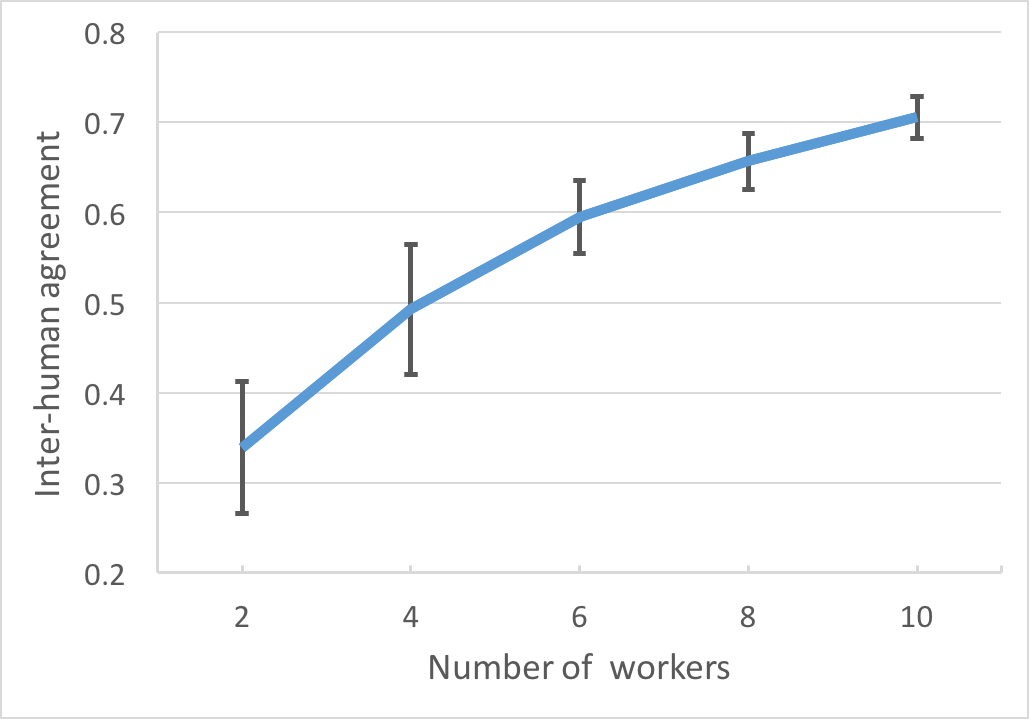}}%
\caption{Inter-human agreement (y-axis) as we collect funniness ratings from more workers (x-axis). We see can see that by 10 ratings, we are starting to saturate with high agreement, indicating that 10 ratings is sufficient for a reliable \emph{funniness score}.}
\label{fig:interhumanAg}%
\end{figure}

Perception of an image differs from one person to another. Moran \etal~\cite{humorAppreciation} treat humor appreciation by people as a personality characteristic. We investigate to what extent people agree how funny each scene in our dataset is. We split the votes we received for each scene into two groups, keeping each individual worker's ratings in the same group to the extent possible. We compute the \emph{funniness score} of each scene across workers in each group. We compute Pearson's correlation between the two groups. \figref{fig:interhumanAg} shows a plot of Pearson's correlation (y-axis) \vs the number of workers (x-axis). We can see that inter-human agreement increases as we increase the number of workers in a group and that the trend is gradually saturating. This indicates that ratings from 10 workers is sufficient to compute a reliable \emph{funniness score}. \par 
We observed that the standard deviation among ratings from 10 different workers for funny scenes is 1.09, and for unfunny scenes is 0.73. \Ie, people agree more on scenes that are clearly not funny than on ones that are funny, matching our intuition that humor is subjective, while the lack thereof is not.  \par
\vspace{10pt}
\noindent
{\fontsize{11}{12}\selectfont \textbf{Appendix II: Object Embeddings}} \par
\label{subsec:embeddings}
\begin{figure*}[t!]
\begin{center}
\setlength{\fboxsep}{0pt}
\setlength{\fboxrule}{0pt}
\fbox{\includegraphics[height=0.45\textwidth, width=0.45\textwidth]{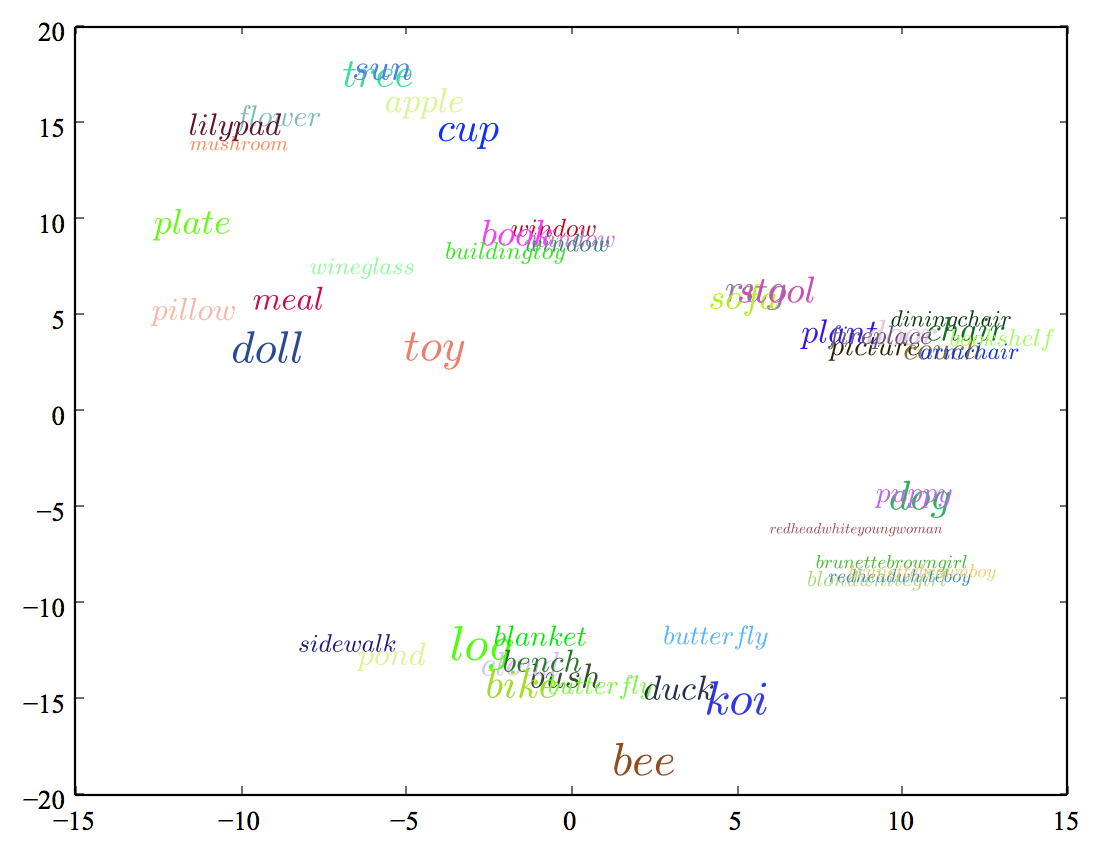}\hspace{0.005\textwidth}}
\fbox{\includegraphics[height=0.45\textwidth, width=0.45\textwidth]{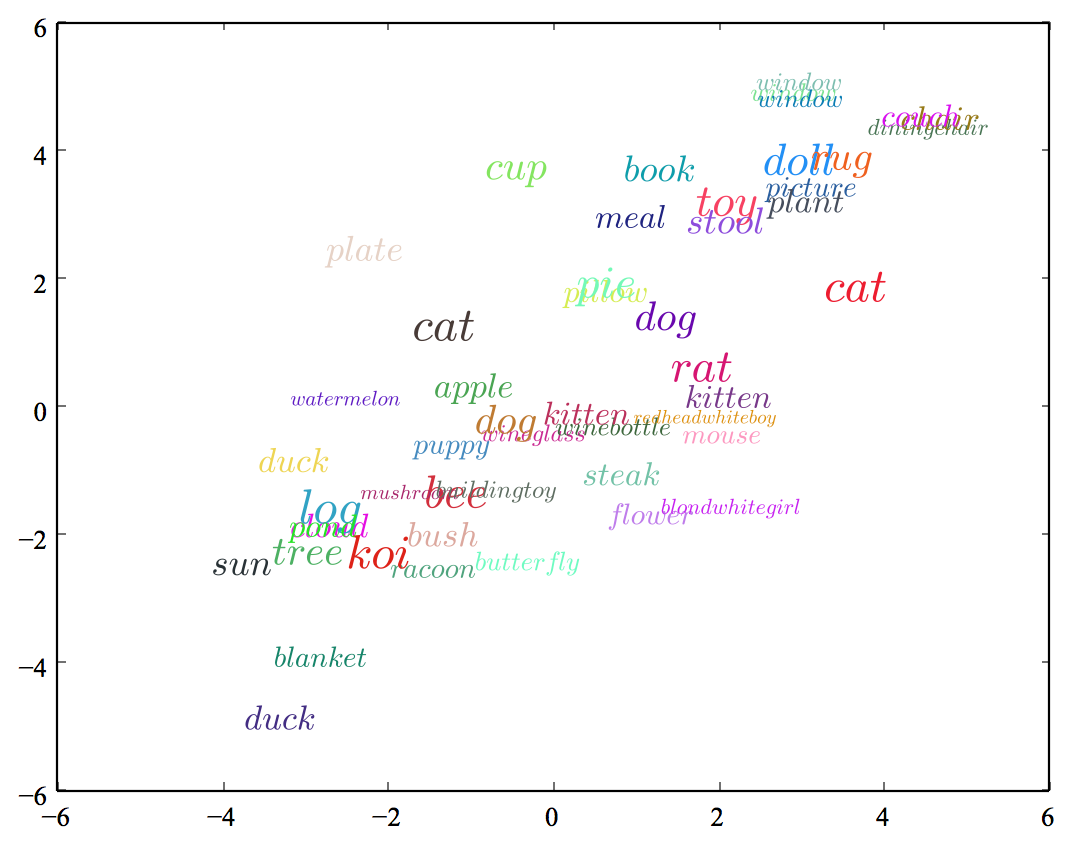}}
\end{center}
\caption{\emph{Left}. Visualization of \quotes{normal} object embeddings of 75 most frequent objects in unfunny scenes. We see that closely placed objects have semantically similar meanings. \emph{Right}. Visualization of \quotes{humor} embeddings of 75 most frequent objects in funny scenes. We see that objects that are close in the \quotes{humor} embedding space may be semantically very different.}
\label{fig:embeddings}
\end{figure*}
\begin{figure}
\begin{center}
\setlength{\fboxsep}{0pt}
\setlength{\fboxrule}{0pt}
\fbox{\includegraphics[height=0.2\textwidth, width=0.45\textwidth]{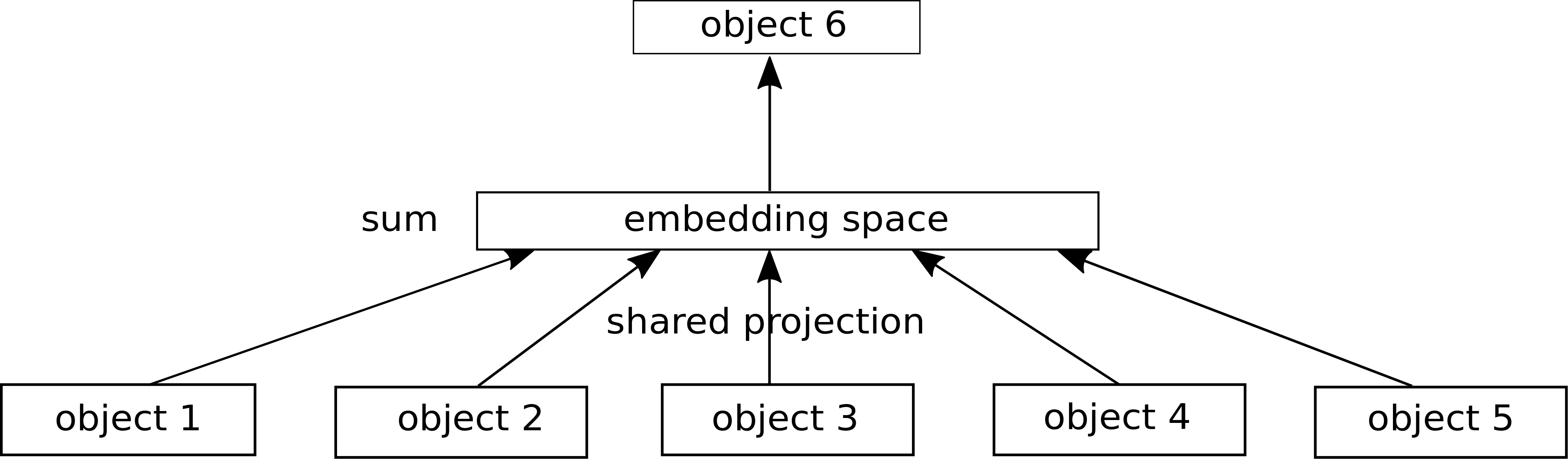}\hspace{0.005\textwidth}}
\caption{The continuous Bag-of-Words model used to obtain the object embeddings.}
\label{fig:embed_model}
\end{center}
\vspace{-10pt}
\end{figure}
\vspace{5pt}
In this section, we describe our model that learns embeddings for clipart objects and present visualizations of these embeddings. We learn distributed representations for each object category in the abstract scenes vocabulary using a word2vec-style continuous Bag-of-Words model~\cite{word2vec}. During training, subsets of 6 objects are sampled from all of the objects present in a scene and the model tries to predict one of the objects, given the other 5. Each object is assigned a 150-d vector, which is randomly initialized. 
The vectors corresponding to the 5 context objects
are projected to an embedding space via a single layer whose parameters are shared between the 5 objects. 
This (randomly initialized) layer consists of 150 hidden units without a non-linearity after it.
The sum of these 5 object projections is used to compute a softmax over the 150 classes in the object vocabulary. 
Using the correct label (\ie, the object category of the 6th object), the cross-entropy loss is computed and backpropagated to learn all network parameters.
The model is trained using Stochastic Gradient Descent with a base learning rate of 0.0001 and a momentum update of 0.9. The learning rate was reduced by a factor of two after each epoch.
A diagram of the model can be seen in \figref{fig:embed_model}. \par
The context provided by the 5 objects ensures that the representations learnt reflect the relationships between objects. \Ie, objects that are semantically related tend to have similar representations. We learn the \quotes{normal} embeddings (\ie, the object embedding instance-level features from the main paper) from 11K scenes collected by Antol~\etal~\cite{vqa}. As these scenes were not intended to be humorous, the relationships captured in the embeddings are the ones that occur naturally in the abstract scenes world. \par
\figref{fig:embeddings} \emph{(left)} is a t-SNE~\cite{tsne} visualization of the \quotes{normal} embeddings for the 75 most frequent objects in unfunny scenes. In~\figref{fig:embeddings} \emph{(right)}, we also visualize \quotes{humor} embeddings, which were not used as features but provide us with insights. These are learnt from the 3,028 funny scenes in the AVH dataset. \par
We observe that the \quotes{normal} embeddings encode a notion for which object categories occur in similar contexts. We also observe that closely placed objects in the \quotes{normal} embedding space have semantically similar meanings. For instance, humans are clustered together around coordinates \emph{(10, -7)}. Interestingly, \quotes{dog} and \quotes{puppy} (coordinates \emph{(10, -5)}) are placed together and furniture like \quotes{chair}, \quotes{bookshelf}, \quotes{armchair}, ~\etc are placed together (coordinates \emph{(10, 5)}). This follows from the distributional hypothesis, which states that words which occur in the similar contexts tend to have similar meanings \cite{wordContextSemantics, distributionalStructure}. \par
\begin{figure*}[t]
\setlength{\fboxsep}{0pt}%
\setlength{\fboxrule}{0pt}%
\begin{center}
\fbox{\includegraphics[width=1\textwidth, height=0.32\textwidth]{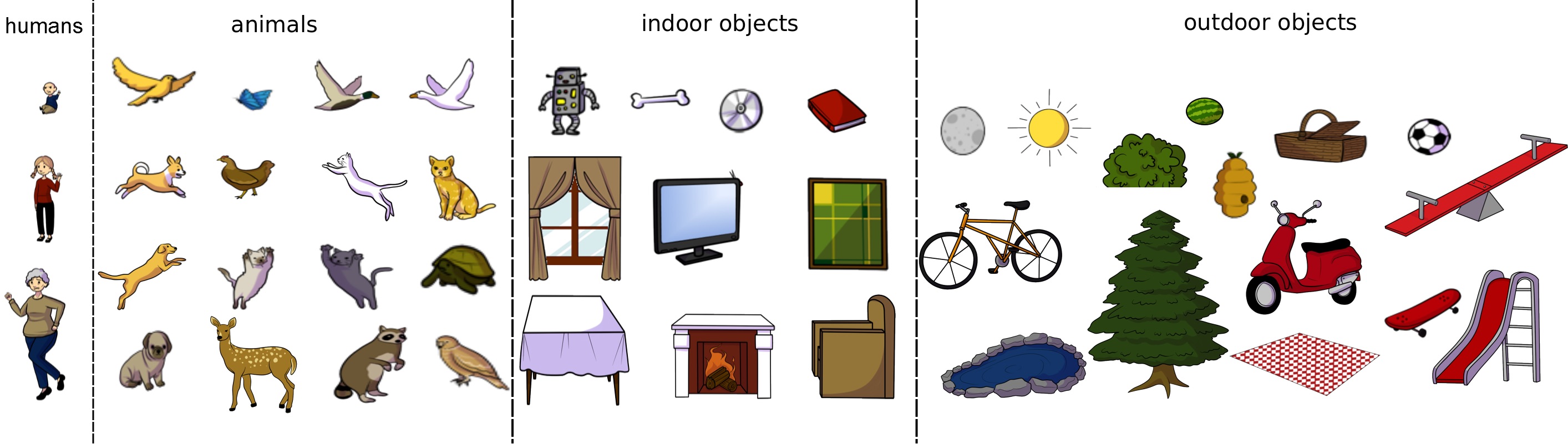}\hspace{0.005\textwidth}}
\end{center}
\caption{A subset of clipart objects from the abstract scenes vocabulary.}%
\label{fig:abstractVocab}%
\vspace{10pt}
\end{figure*}%
\noindent
\begin{figure*}[t!]
\setlength{\fboxsep}{0pt}
\setlength{\fboxrule}{0pt}
\centering
	\begin{subfigure}[t]{1.5in}
		\centering
		\fbox{\includegraphics[width=1.65in]{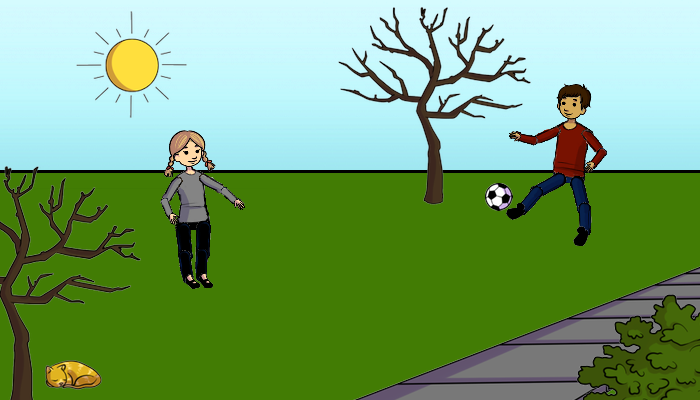}}
		\caption{1.3}\label{fig:boring}
	\end{subfigure}
	\quad
	\begin{subfigure}[t]{1.5in}
		\centering
		\fbox{\includegraphics[width=1.65in]{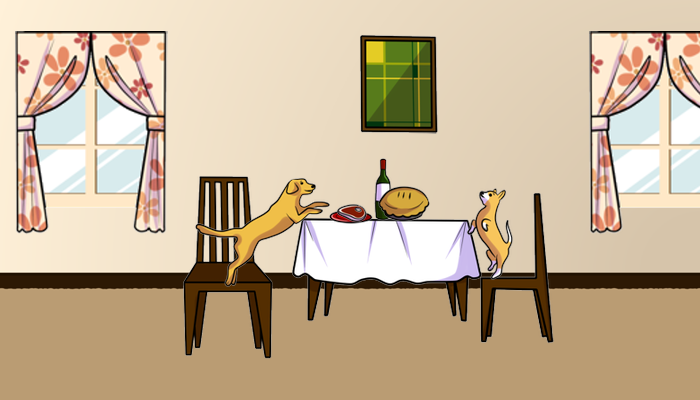}}
		\caption{2.8}\label{fig:dogDinner}
	\end{subfigure}
	\quad
	\begin{subfigure}[t]{1.5in}
		\centering
		\fbox{\includegraphics[width=1.65in]{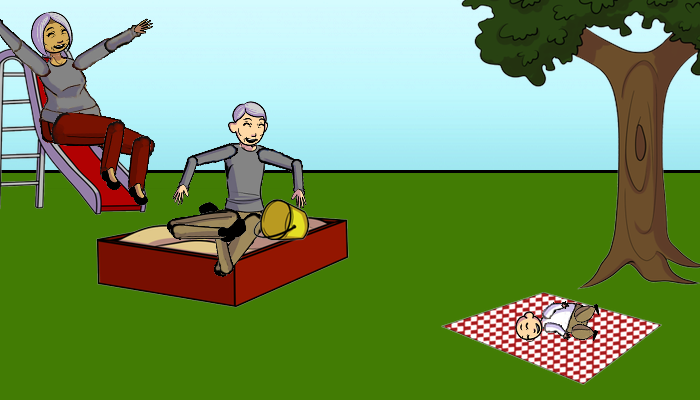}}
		\caption{3.2}\label{fig:oldPeopleFun}
	\end{subfigure}
	\quad
	\begin{subfigure}[t]{1.5in}
		\centering
		\fbox{\includegraphics[width=1.65in]{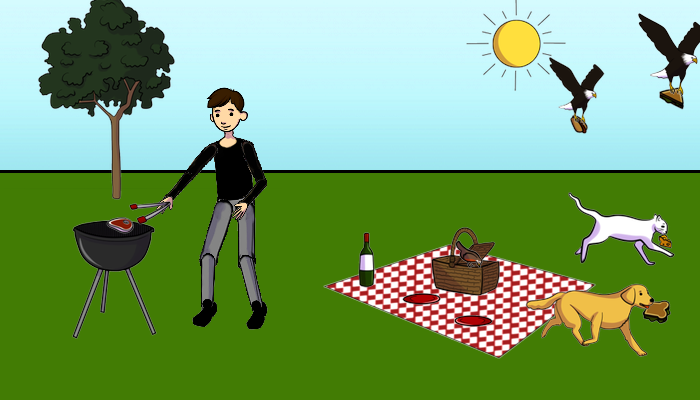}}
		\caption{4.4}\label{fig:animalStealing}
	\end{subfigure}
	\quad
	\begin{subfigure}[t]{1.5in}
		\centering
		\fbox{\includegraphics[width=1.65in]{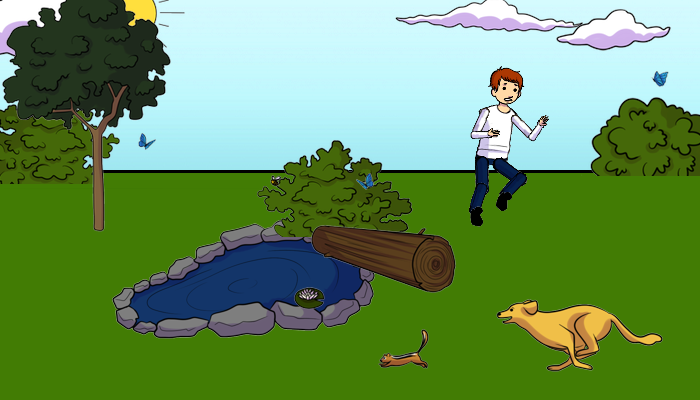}}
		\caption{1.1}\label{fig:normalPark}
	\end{subfigure}
	\quad
	\begin{subfigure}[t]{1.5in}
		\centering
		\fbox{\includegraphics[width=1.65in]{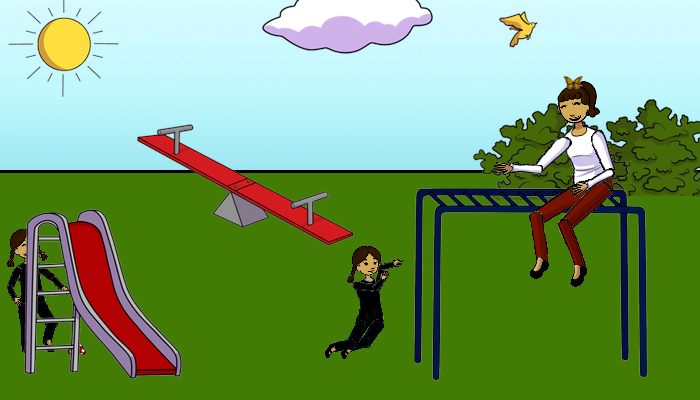}}
		\caption{2.7}\label{fig:ladyBars}
	\end{subfigure}
	\quad
	\begin{subfigure}[t]{1.5in}
		\centering
		\fbox{\includegraphics[width=1.65in]{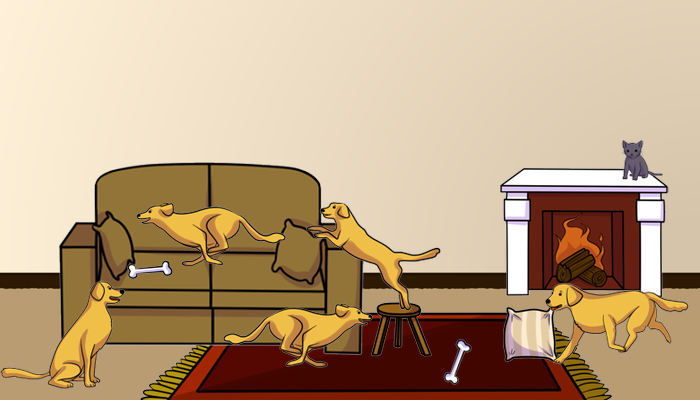}}
		\caption{3.5}\label{fig:dogsEverywhere}
	\end{subfigure}
	\quad
	\begin{subfigure}[t]{1.5in}
		\centering
		\fbox{\includegraphics[width=1.65in]{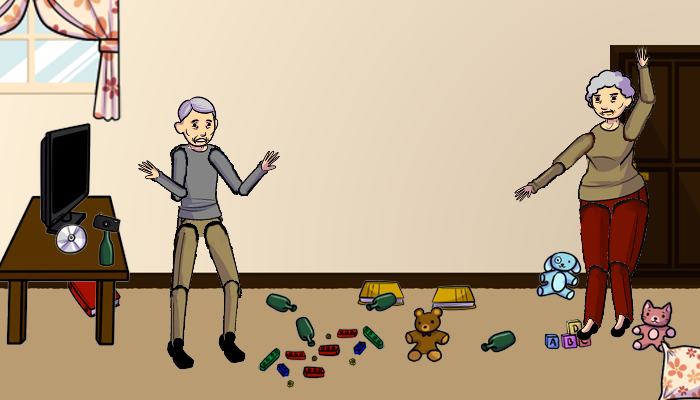}}
		\caption{4.1}\label{fig:messyGrandpa}
	\end{subfigure}
	\quad
\caption{Spectrum of scenes from our AVH dataset that are arranged in ascending order of \emph{funniness 
score} (shown in the sub-caption)}
\label{fig:exampleScenesAVH}
\end{figure*} 
In contrast, in the \quotes{humor} embeddings, visualized in \figref{fig:embeddings} \emph{(right)}, we see that objects that are close in the embedding space may be semantically very different. For instance, \quotes{dog} and \quotes{wine glass} are placed together at coordinates \emph{(0, 0)}. These are placed far apart (at opposite ends) in the \quotes{normal} embedding. However, in the \quotes{humor} embedding, these two categories are extremely close to each other; even closer than semantically similar categories like two breeds of dogs. We hypothesize that this because our dataset contains funny scenes consisting of dogs with wine glasses, \eg,~\figref{fig:dogDinner}. It is interesting to note that \quotes{background} objects that do not contribute to humor in a scene are also placed together. For example, \quotes{chair}, \quotes{couch}, and \quotes{window} are placed together in the \quotes{humor} embedding as well (coordinates \emph{(4, 5)}).  \par
The understanding of semantically similar object categories that can occur in a context, represented by the \quotes{normal} embeddings, can be interpreted as a person's mental model of the world. The \quotes{humor} embeddings capture deviations or incongruities from this \quotes{normal} view that might cause humor.  \par
\vspace{10pt} 
\noindent
\begin{figure}[t]
\setlength{\fboxsep}{0pt}%
\setlength{\fboxrule}{0pt}%
		\fbox{\includegraphics[width=0.235\textwidth]{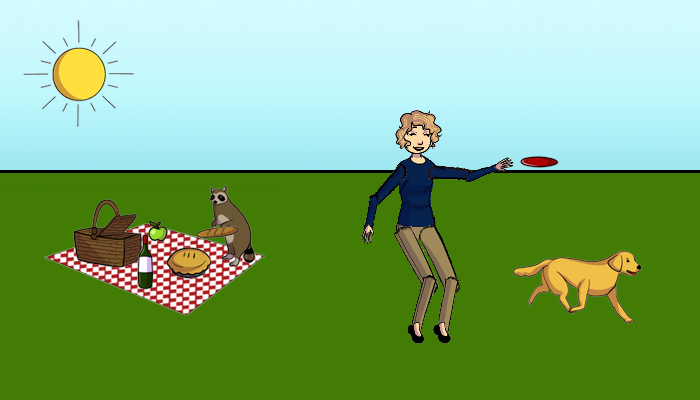}\hspace{0.00005\textwidth}}
		\fbox{\includegraphics[width=0.235\textwidth]{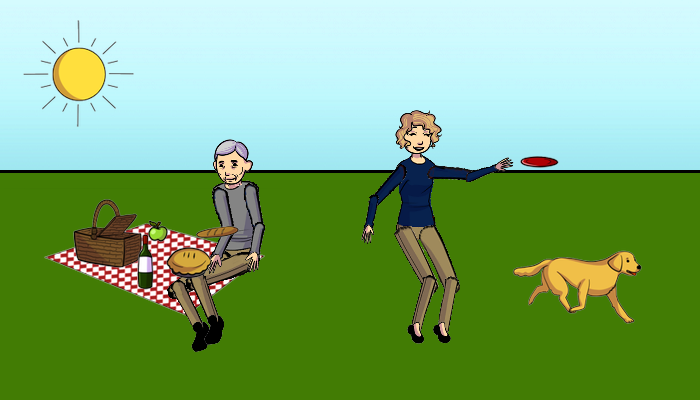}}\vspace{0.00005\textwidth}
		\fbox{\includegraphics[width=0.235\textwidth]{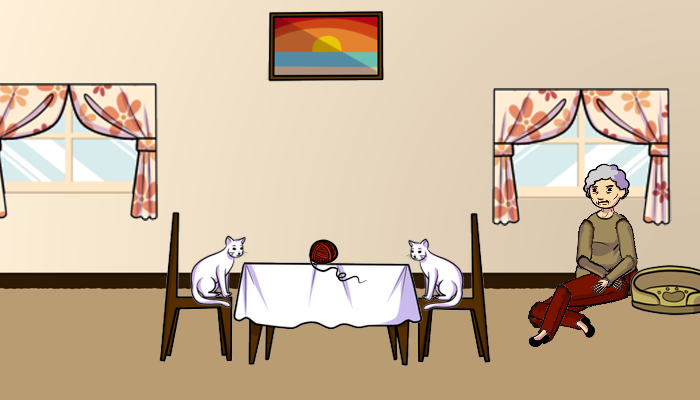}\hspace{0.00005\textwidth}}
		\fbox{\includegraphics[width=0.235\textwidth]{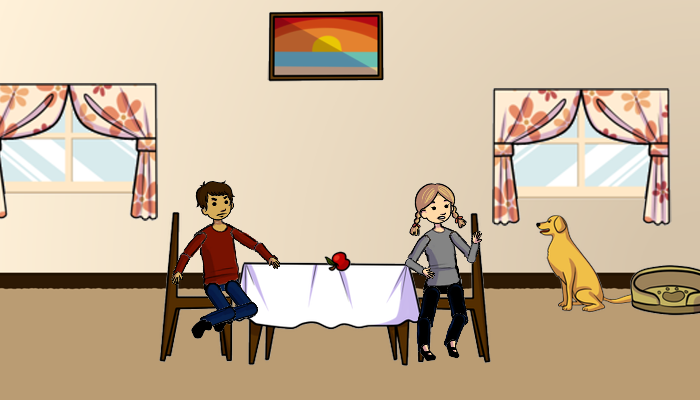}}\vspace{0.00005\textwidth}
		\fbox{\includegraphics[width=0.235\textwidth]{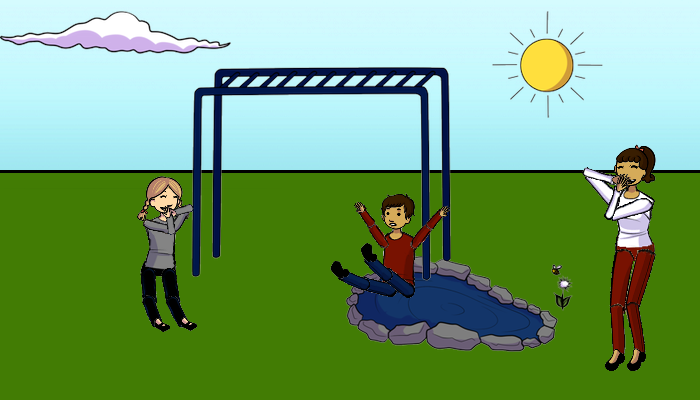}\hspace{0.00005\textwidth}}
		\fbox{\includegraphics[width=0.235\textwidth]{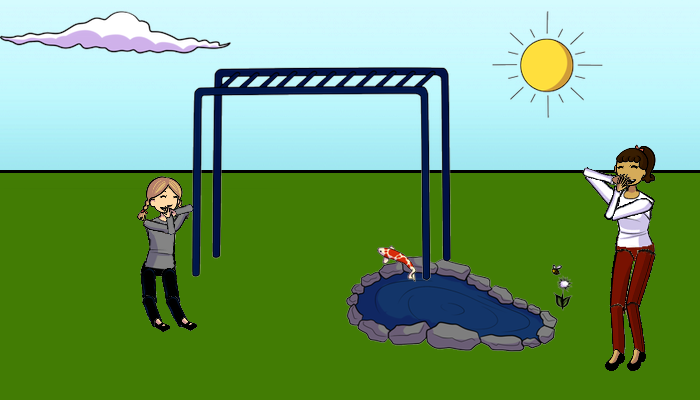}}\vspace{0.00005\textwidth}
		\fbox{\includegraphics[width=0.235\textwidth]{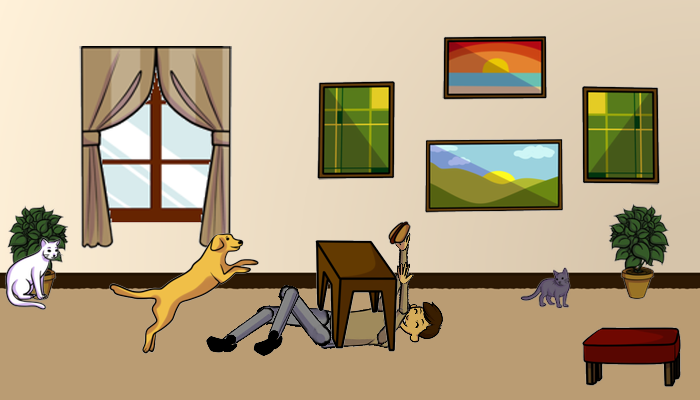}\hspace{0.00005\textwidth}}
		\fbox{\includegraphics[width=0.235\textwidth]{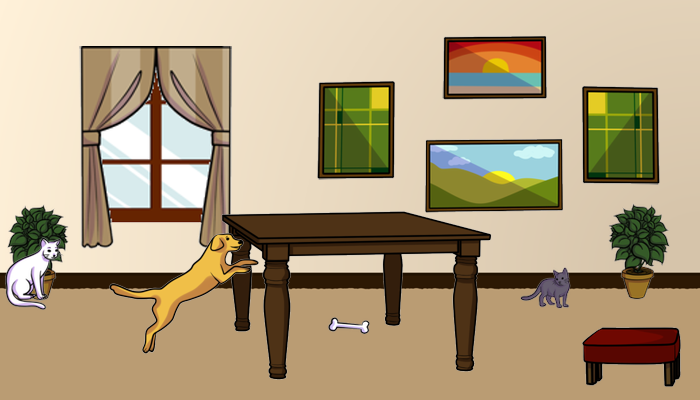}}\vspace{0.00005\textwidth}
		\fbox{\includegraphics[width=0.235\textwidth]{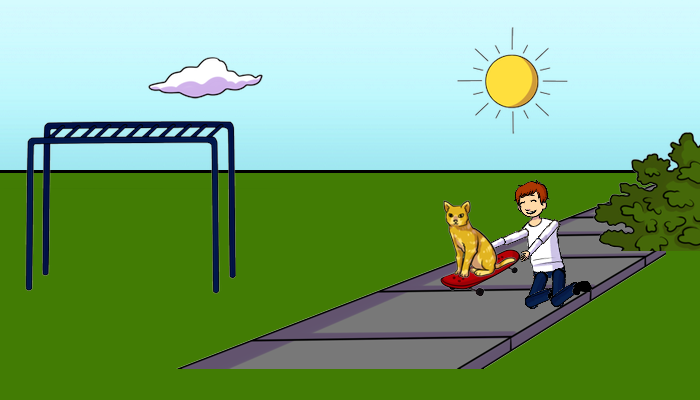}\hspace{0.00005\textwidth}}
		\fbox{\includegraphics[width=0.235\textwidth]{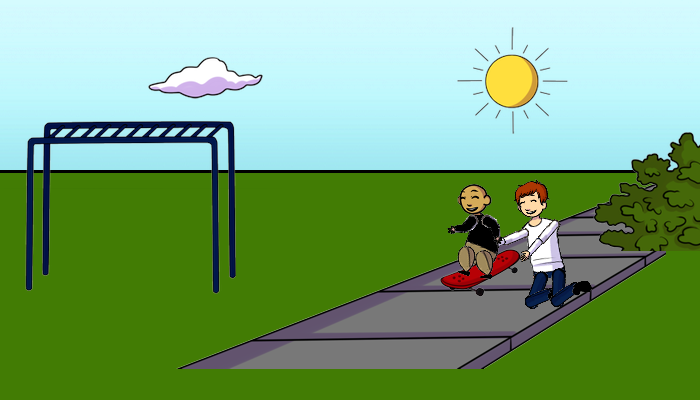}}
\caption{Some example originally funny scenes (\emph{left}) and their object-replaced unfunny counterparts (\emph{right}) from the FOR dataset.}
\label{fig:exampleScenesFOR}
\end{figure}
{\fontsize{12}{12}\selectfont \textbf{Appendix III: Abstract Scenes Vocabulary}} \par
\label{sec:abstractVocab}
\vspace{5pt}
The abstract scenes interface developed by Antol~\etal~\cite{vqa} consists of 20 \quotes{deformable} humans, 31 animals in different poses, and about 100 objects that can be found in indoor scenes (\eg, couch, picture, doll, door, window, plant, fireplace) or outdoor scenes (\eg, tree, pond, sun, clouds, bench, bike, campfire, grill, skateboard). In addition to the 8 different expressions available for humans, the ability to vary the pose of a human at a fine-grained level enables these abstract scenes to effectively capture the semantics of a scene. The large clipart vocabulary (of which only a fraction is shown to a worker during creation of a scene) ensures diversity in the scenes being depicted. A subset of objects from our Abstract Scenes vocabulary is shown in \figref{fig:abstractVocab}. \newline \newline \par
\noindent
{\fontsize{12}{12}\selectfont \textbf{Appendix IV: Example Scenes}}  \par
\begin{figure*}[t]
\setlength{\fboxsep}{0pt}%
\setlength{\fboxrule}{0pt}%
\begin{center}
\fbox{\includegraphics[width=1\textwidth]{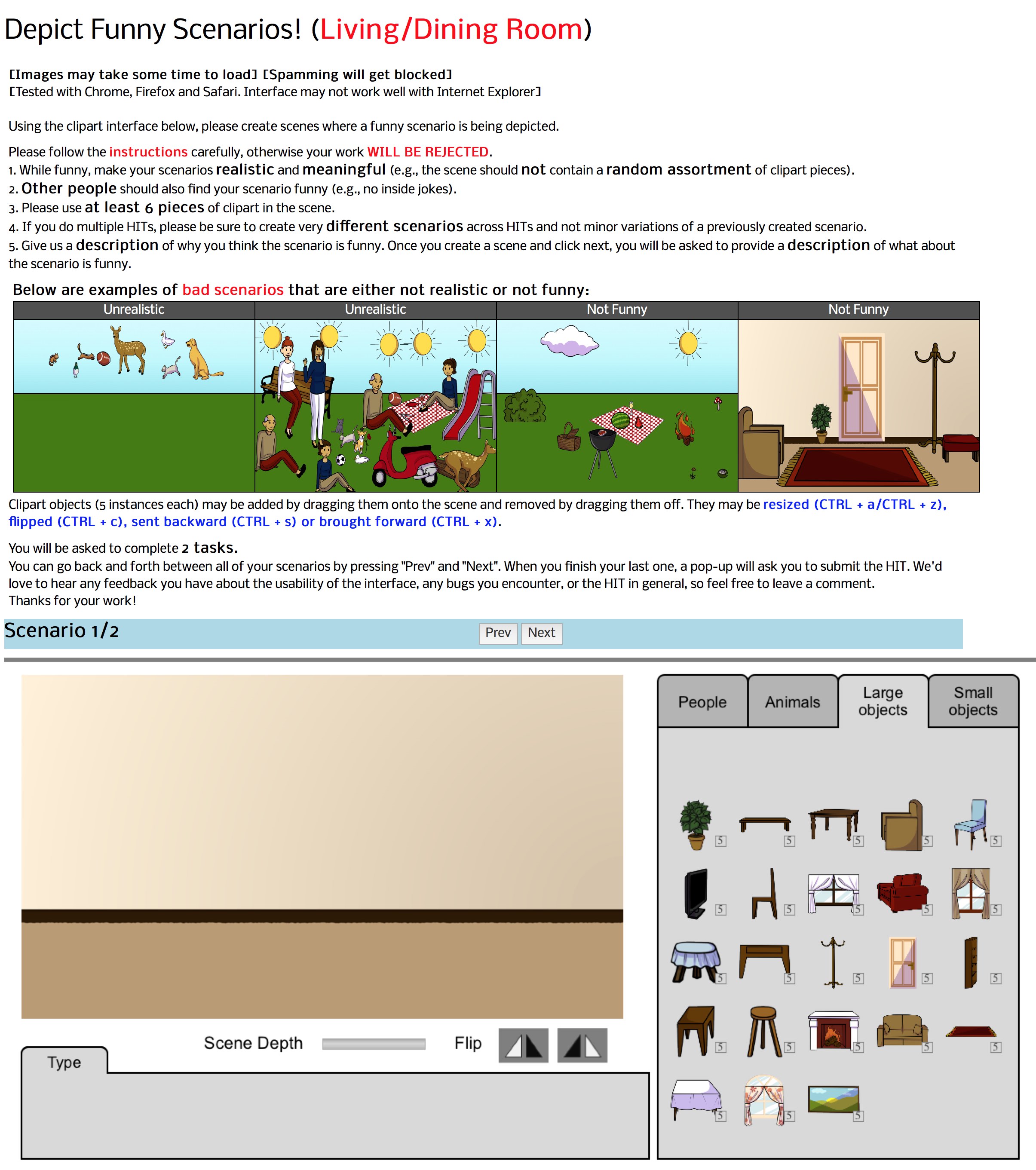}\hspace{0.005\textwidth}}
\end{center}
\caption{User interface used to create the funny scenes in the AVH dataset.}%
\label{fig:createFunny}%
\end{figure*}

\label{sec:exampleScenes}
\vspace{5pt}
In this section, we present examples of scenes that were created using the abstract scenes interface. \figref{fig:exampleScenesAVH}, depicts a spectrum of scenes from the AVH dataset in ascending order of \emph{funniness score}. These scenes were created by AMT workers using the interface presented in~\figref{fig:createFunny}. \par
\begin{figure*}[t]
\setlength{\fboxsep}{0pt}%
\setlength{\fboxrule}{0pt}%
\fbox{\includegraphics[width=0.8\textwidth]{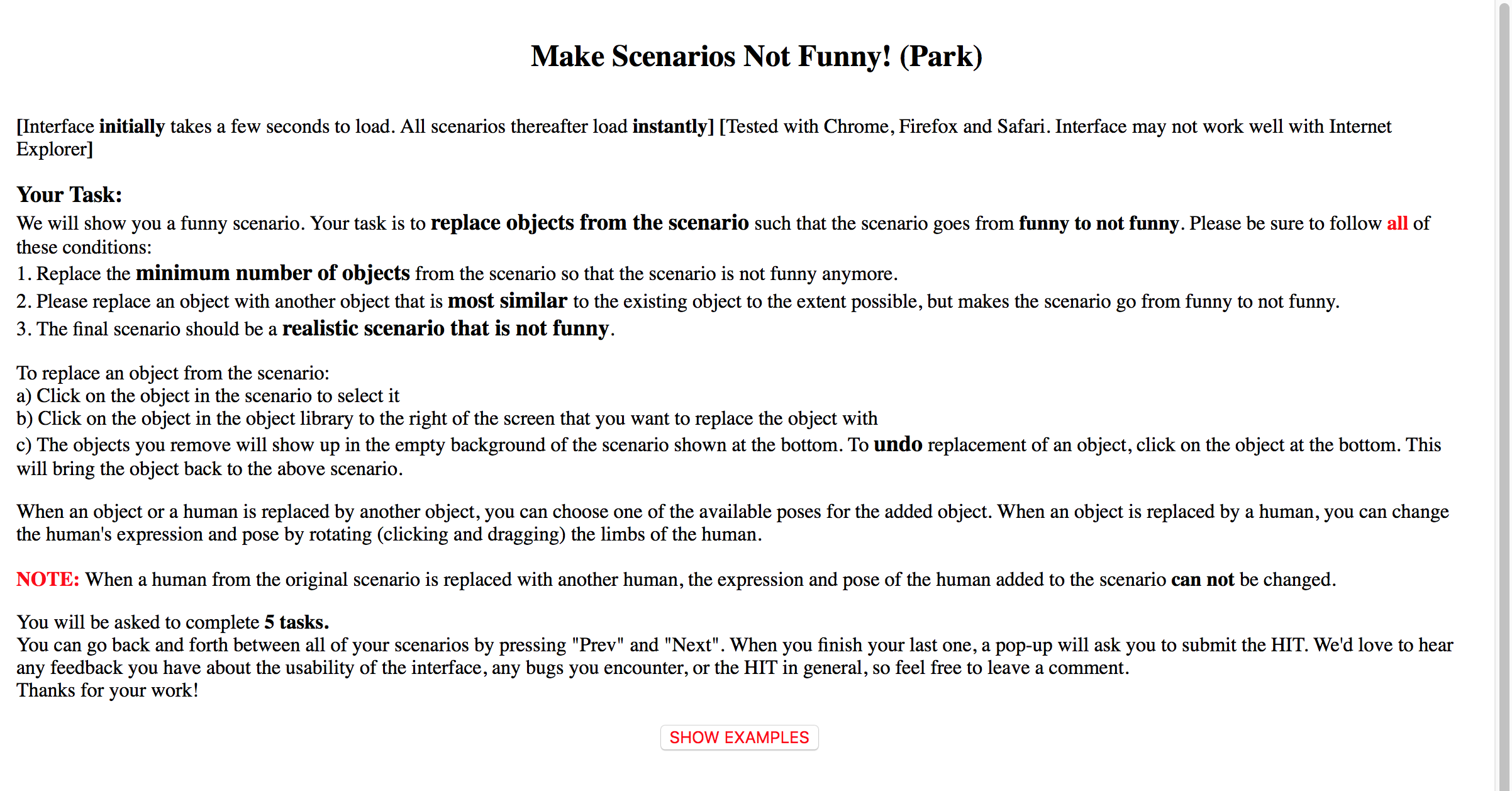}}\vspace{0pt}
\fbox{\includegraphics[width=0.8\textwidth]{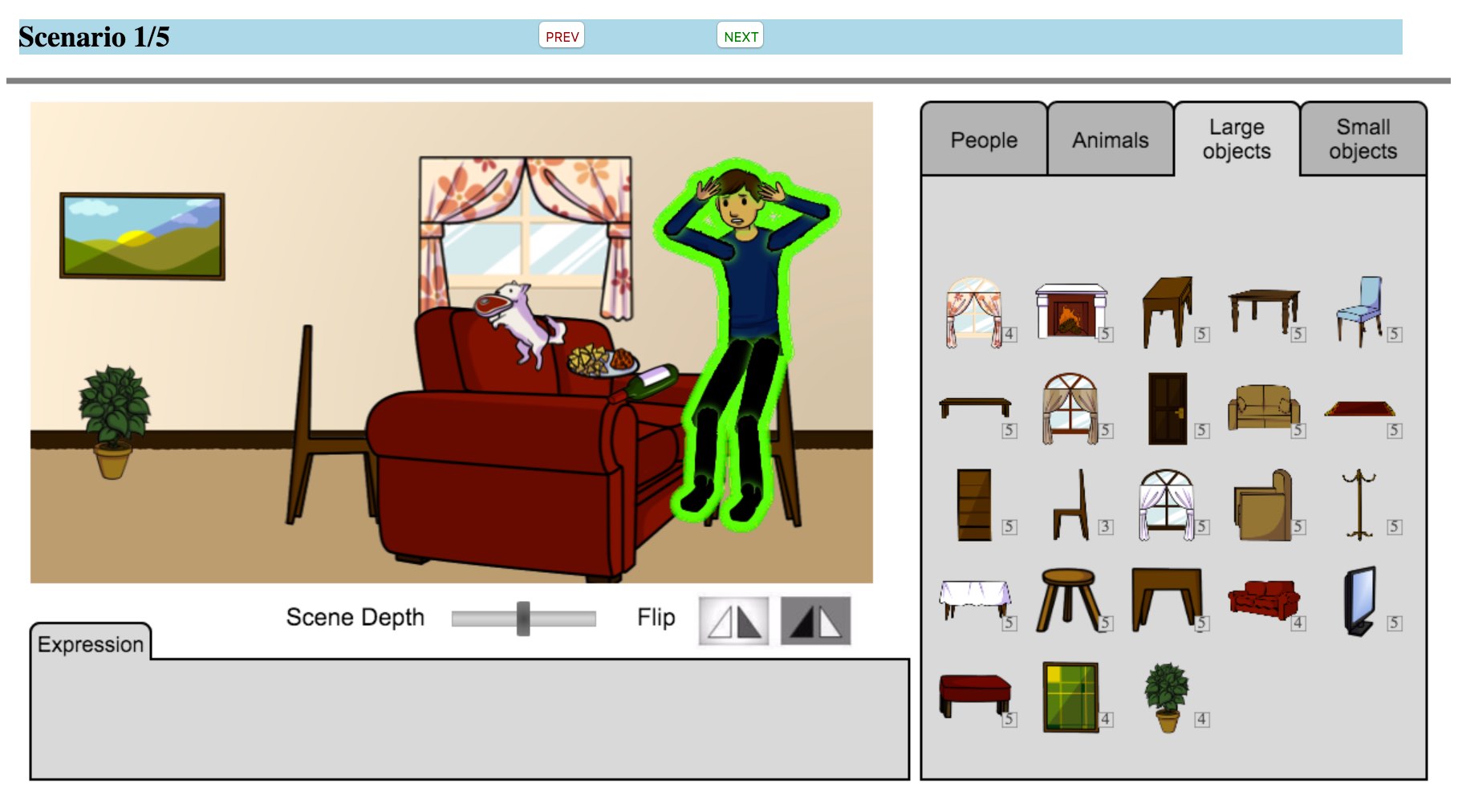}}\vspace{0pt}
\fbox{\includegraphics[width=0.8\textwidth]{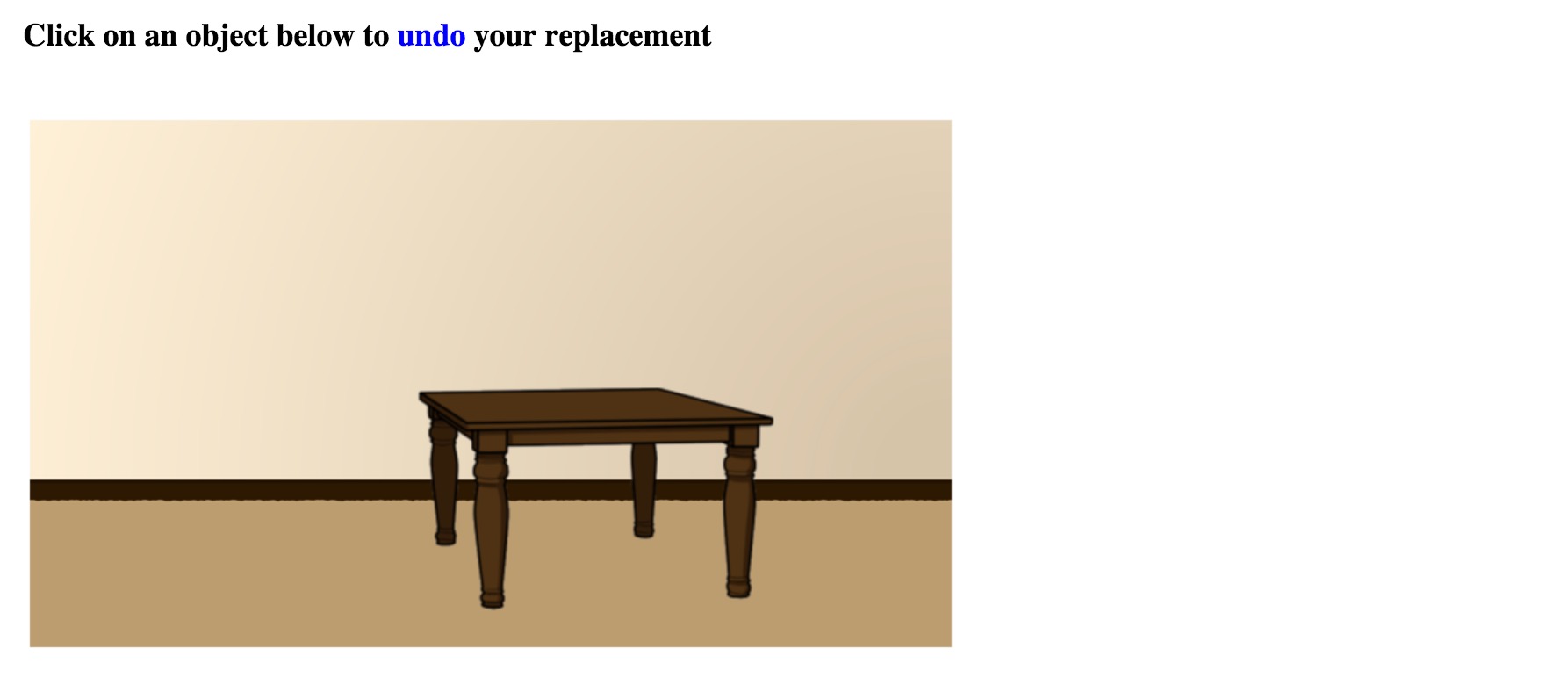}}
\caption{User interface to replace objects for the FOR dataset.}
\label{fig:replaceObjects}%
\end{figure*}

\figref{fig:exampleScenesFOR} shows originally funny scenes (\emph{left}) and their unfunny counterparts (\emph{right}) from the FOR dataset. AMT workers created the counterparts by replacing as few objects in the originally funny scene such that the resulting scene is not funny anymore. A screenshot of the interface that was used to create the unfunny counterparts is shown in~\figref{fig:replaceObjects}. \newline \par
\noindent
{\fontsize{12}{12}\selectfont \textbf{Appendix V: Object Type Occurrences}}  \par
\label{sec:distObjs}
\vspace{5pt}
In this section, we first analyze the occurrence of each object type in funny and unfunny scenes. We then analyze the most commonly cooccurring object types in funny scenes as compared to unfunny scenes. \par
\noindent
\begin{figure}
\setlength{\fboxsep}{0pt}%
\setlength{\fboxrule}{0pt}%
\center
\fbox{\includegraphics[width=0.45\textwidth]{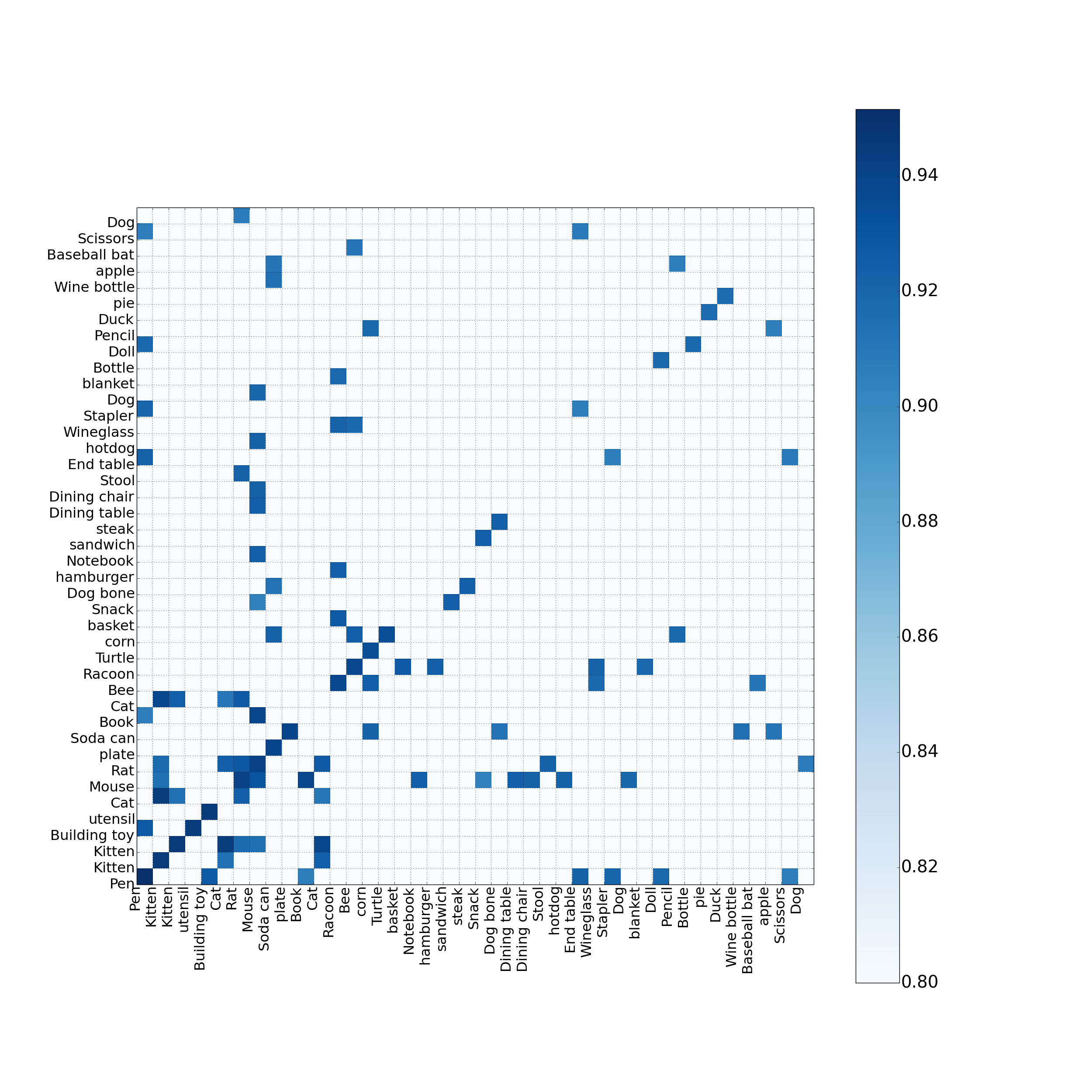}}
\caption{Top 100 object pairs that have the highest probabilities of cooccurring in a funny scene. Please note that repeated entries for an object type (\eg, \quotes{dog}), correspond to slightly different versions (\eg, breeds) of the same object type.}
\label{fig:coOccur}%
\end{figure}
\begin{figure}[t]
\vspace{-3pt}
\setlength{\fboxsep}{0pt}%
\setlength{\fboxrule}{0pt}%
\begin{center}
\fbox{\includegraphics[width=0.49\textwidth]{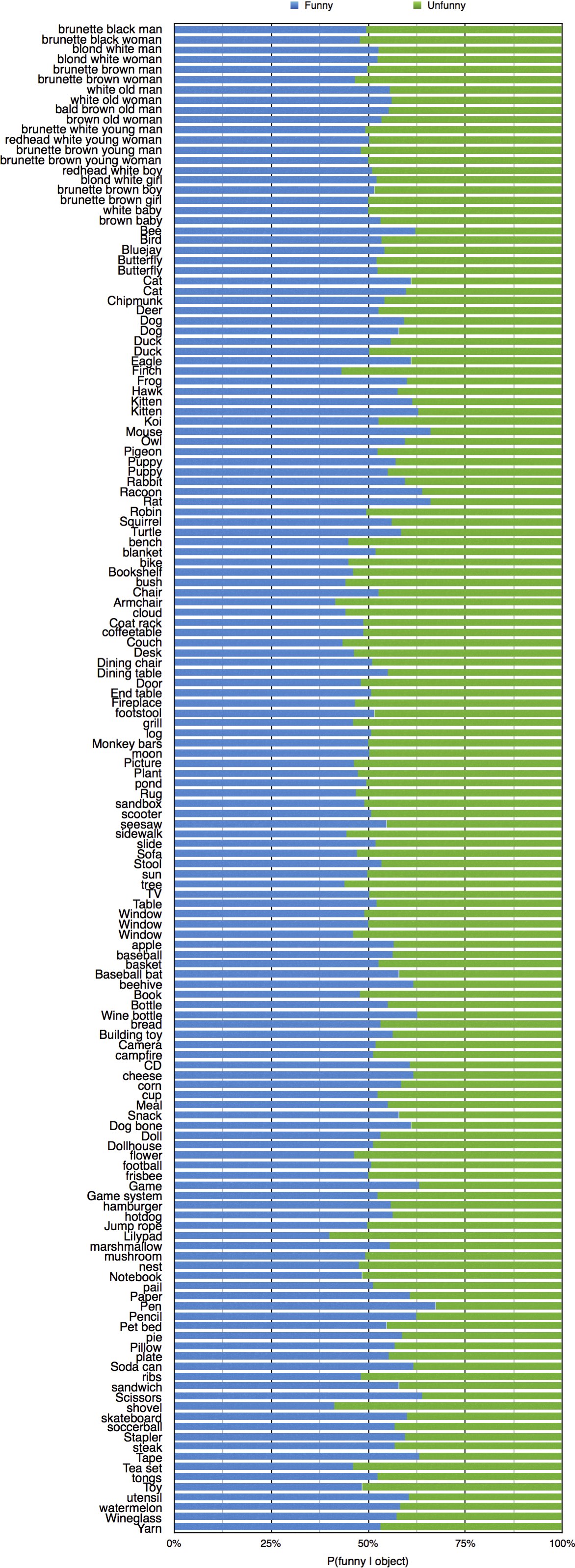}}\vspace{0pt}
\caption{Probability of scene being funny, given object.}
\label{fig:probFunny}%
\end{center}
\end{figure}

\noindent
\textbf{Distribution of Object Types.} We analyze the distribution of object types in funny and unfunny scenes across all scenes in our dataset. We compute the frequency of appearance of each object type in funny and unfunny scenes. We use this to compute the probability of a scene being funny, given that an object is present in the scene, which is shown in \emph{blue} in~\figref{fig:probFunny}. Since we have more unfunny scenes than funny scenes, we use normalized counts. \par
We observe that the humans that most appear in funny scenes are elderly people. This is probably because a number of scenes in our dataset depict old men behaving unexpectedly,~\eg., dancing or playing in the park as shown in~\figref{fig:oldPeopleFun}, which is funny.
Interestingly, we also observe that in general, animals appear more frequently in funny scenes. Animals like \quotes{mouse}, \quotes{rat}, \quotes{raccoon} and \quotes{bee} appear in funny scenes significantly more than they do in unfunny scenes. Other objects having a strong bias towards appearing in funny scenes include \quotes{wine bottle}, \quotes{pen}, \quotes{scissors}, \quotes{tape}, \quotes{game} and \quotes{beehive}. Thus, we see that certain object types have a tendency to appear in funny scenes. A possible reason for this is that these objects are involved in funny interactions, or are intrinsically funny, and hence contribute to humor in these scenes. \newline
\textbf{Funny Cooccurrence Matrix.} We populate two object cooccurrence matrices -- \textbf{F} and \textbf{U}, corresponding to funny scenes and unfunny scenes, respectively. Each element in \textbf{F} and \textbf{U} corresponds to the count of the cooccurrence of a pair of objects across all funny and unfunny scenes, respectively. To enable the study of types of cooccurrences that contribute to humor, we compute the probability of a scene being funny, given that a pair of objects cooccur in the scene as $\frac{\textbf{F}}{\textbf{F}+\textbf{U}}$, which is shown in \figref{fig:coOccur} for the top 100 probable combinations that exist in a funny scene. Please note that repeated entries for an object type (\eg, \quotes{dog}), correspond to slightly different versions (\eg, breeds) of the same object type. An interesting set of object pairs that are present in funny scenes are \quotes{rat} appearing alongside \quotes{kitten}, \quotes{cat}, \quotes{stool}, and \quotes{dog}. Another interesting set of combinations is \quotes{raccoon} cooccurring with \quotes{bee}, \quotes{hamburger}, \quotes{basket}, and \quotes{wine glass}. We observe that this matrix captures interesting and unusual combinations of objects that appear together frequently in funny scenes. \newline \par
\vspace{10pt}
\noindent
{\fontsize{12}{12}\selectfont \textbf{Appendix VI: User Interfaces}}  \par
\label{sec:interfaces}
\vspace{5pt}
In this section, we present the user interfaces that were used to collect data from AMT. \figref{fig:createFunny} shows a screenshot of the user interface that we used to collect funny scenes. Objects in the clipart library (on the \emph{right} in the screenshot) can be dragged on to any part of the empty canvas shown in the figure. The pose, flip (\ie, lateral orientation), and size of all objects can be changed once they are placed in the scene. In the case of humans, one of 8 expressions must be chosen (initially humans have blank faces) and fine-grained pose adjustments are required. \par
\figref{fig:replaceObjects} shows the interface that we used to collect \quotes{object-replaced} scenes for our FOR dataset. We showed workers an originally funny scene and asked them to replace objects in that scene so that the scene is not funny anymore. On clicking an object in the original scene, the object gets highlighted in green. A replacer object can then be chosen from the clipart library (displayed on the \emph{right} in the screenshot). Objects that are replaced in the original scene show up in the empty canvas below. At any point, to undo a replacement, a user can click on the object in the below canvas and the corresponding object will be placed at its original position in the scene. The interface does not allow for the movement or the removal of objects. 
\noindent
\noindent
\clearpage %
{\footnotesize
\bibliographystyle{ieee}
\bibliography{refs}
}

{\footnotesize
\bibliographystyle{ieee}
\bibliography{refs}
}
\end{document}